\pdfoutput=1
\documentclass[runningheads]{llncs}
\usepackage{xspace}
\makeatletter
\DeclareRobustCommand\onedot{\futurelet\@let@token\@onedot}
\def\@onedot{\ifx\@let@token.\else.\null\fi\xspace}

\def\eg{\emph{e.g}\onedot} 
\def\ie{\emph{i.e}\onedot} 
\def\cf{\emph{cf}\onedot} 
\def\etc{\emph{etc}\onedot} \def\vs{\emph{vs}\onedot}
 
\def\etal{\emph{et al}\onedot}
\makeatother

\renewcommand{\thefootnote}{\fnsymbol{footnote}}

\usepackage{graphicx}
\usepackage{comment}
\usepackage{cite}
\usepackage{epsfig}
\usepackage{algpseudocode}
\usepackage{algorithm}
\usepackage{dsfont}
\usepackage{amsmath,amssymb} 
\usepackage{multirow}

\usepackage{color}
\usepackage{capt-of}
\usepackage{booktabs}
\usepackage{varwidth}

\usepackage[table]{xcolor}

\definecolor{lightblue}{RGB}{225, 225, 255}
\definecolor{lightred}{RGB}{255, 225, 225}

\newcommand{\PAR}[1]{\vskip4pt \noindent{\bf #1~}}

\usepackage[pagebackref=true,breaklinks=true,colorlinks,bookmarks=false]{hyperref}

\begin{document}
\pagestyle{headings}
\mainmatter
\def\ECCVSubNumber{2583}  

\title{
Single-Image Depth Prediction Makes Feature Matching Easier 
}

\titlerunning{Depth for Feature Matching}
\author{
Carl~Toft\textsuperscript{$1^\star$}
\and
\hspace{10pt}
Daniyar~Turmukhambetov\textsuperscript{$2$} 
\and
\hspace{10pt}
Torsten~Sattler\textsuperscript{$1$}
\and
\\\vspace{3pt}\hspace{10pt}
Fredrik~Kahl\textsuperscript{$1$}
\and
\hspace{3pt}
Gabriel~J.~Brostow\textsuperscript{$2,3$} 
}
\authorrunning{C. Toft et al.}
%
\institute{
\textsuperscript{$1$}Chalmers University of Technology \qquad  \textsuperscript{$2$}Niantic \qquad
\textsuperscript{$3$}University College London\\
\vspace{3pt}
\url{www.github.com/nianticlabs/rectified-features}
}

\maketitle

\footnotetext[1]{Work done during an internship at Niantic.}
\renewcommand{\thefootnote}{\arabic{footnote}}

\begin{abstract}
    Good local features improve the robustness of many 3D re-localization and multi-view reconstruction pipelines. The problem is that viewing angle and distance severely impact the recognizability of a local feature. Attempts to improve appearance invariance by choosing better local feature points or by leveraging outside information, have come with pre-requisites that made some of them impractical. In this paper, we propose a surprisingly effective enhancement to local feature extraction, which improves matching. We show that CNN-based depths inferred from single RGB images are quite helpful, despite their flaws. They allow us to pre-warp images and rectify perspective distortions, to significantly enhance SIFT and BRISK features, 
    enabling more good matches, even when cameras are looking at the same scene but in opposite directions. 

\end{abstract}

\section{Introduction}
\noindent Matching local features between images is a core research problem in Computer Vision. 
Feature matching is a crucial step in Simultaneous Localization and Mapping (SLAM)~\cite{Davison07PAMI,MurArtal2017TRO}, Structure-from-Motion (SfM)~\cite{Snavely06SIGGRAPH,schoenberger2016sfm}, and visual localization~\cite{Svarm2017PAMI,Sattler2017PAMI,Sarlin2019CVPR}. By extension, good feature matching enables applications such as self-driving cars~\cite{Heng2019ICRA} and other autonomous robots~\cite{Lim2012CVPR} as well as Augmented, Mixed, and Virtual Reality. 
Handling larger viewpoint changes is often important in practice, \eg, to detect loop closures when revisiting the same place in SLAM~\cite{Galvez2012TRO}  or for re-localization under strong viewpoint changes~\cite{Schoenberger2018CVPR}.

\begin{figure}[t]
    \centering
    \includegraphics[width=0.3\linewidth]{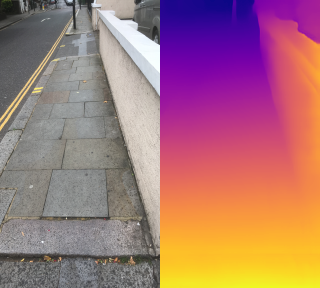} \hspace{1pt} \includegraphics[width=0.3\linewidth]{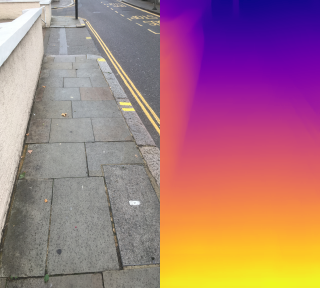} \hspace{1pt}
    \includegraphics[width=0.31\linewidth]{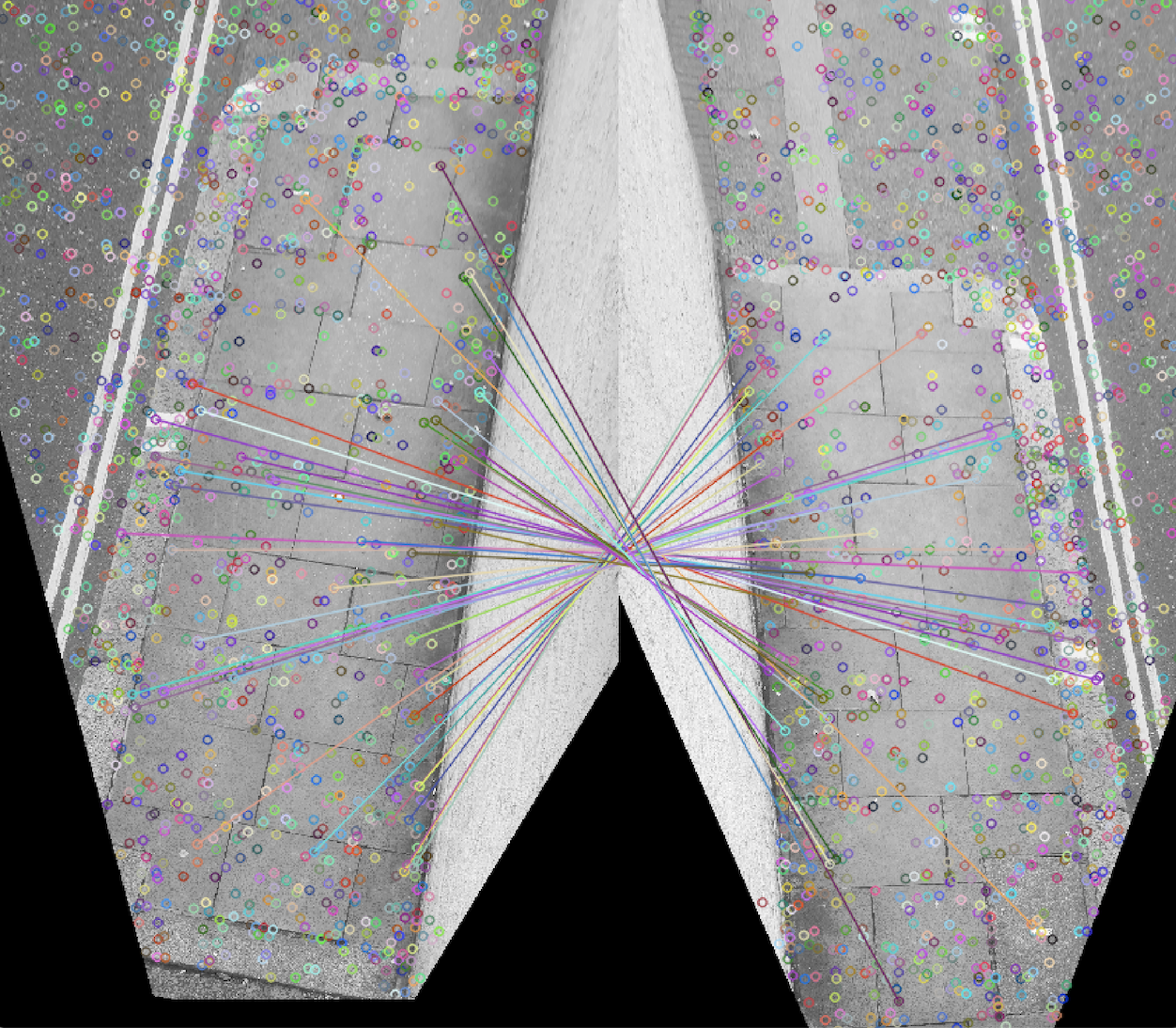}
    \caption{Left to Right: Two input images, capturing the same section of a sidewalk, but looking in opposite directions. Each RGB input image is shown together with its depth map predicted by a single-image depth prediction network. Based on the predicted depth map, we identify planar regions and remove perspective distortion from them before extracting local features, thus enabling effective feature matching under strong viewpoint changes.}
    \label{fig:teaser}
\end{figure}

Traditionally, local features are computed in two stages~\cite{lowe2004distinctive}: the feature detection stage determines salient points in an image around which patches are extracted. 
The feature description stage computes descriptors from these patches. 
Before extracting a patch, the feature detector typically accounts for certain geometric transformations, thus making the local features robust or even invariant against these transformations. 
For example, aligning a patch to a dominant direction makes the feature invariant to in-plane rotations~\cite{lowe2004distinctive,Yi2016CVPR}; detecting salient points at multiple scales introduces robustness to scale changes~\cite{Lindeberg1994JAS}; removing the effect of affine transformations~\cite{Mikolajczyk2005IJCV,Matas2004IMAVIS} of the image makes the extracted features more robust against viewpoint changes~\cite{Aanaes2012IJCV,Fraundorfer2005CVPRW}. 
If the 3D geometry of the scene is known, \eg, from 3D reconstruction via SfM, it is possible to undo the effect of perspective projection before feature extraction~\cite{wu20083d,Zeisl2013ICCV,Zeisl2012ECCVW,Koeser2007ICCV}. The resulting features are, in theory, invariant to viewpoint changes. 

\begin{figure}[t]
    \centering
\begin{minipage}{0.32\textwidth}\centering
    \includegraphics[width=0.6\linewidth]{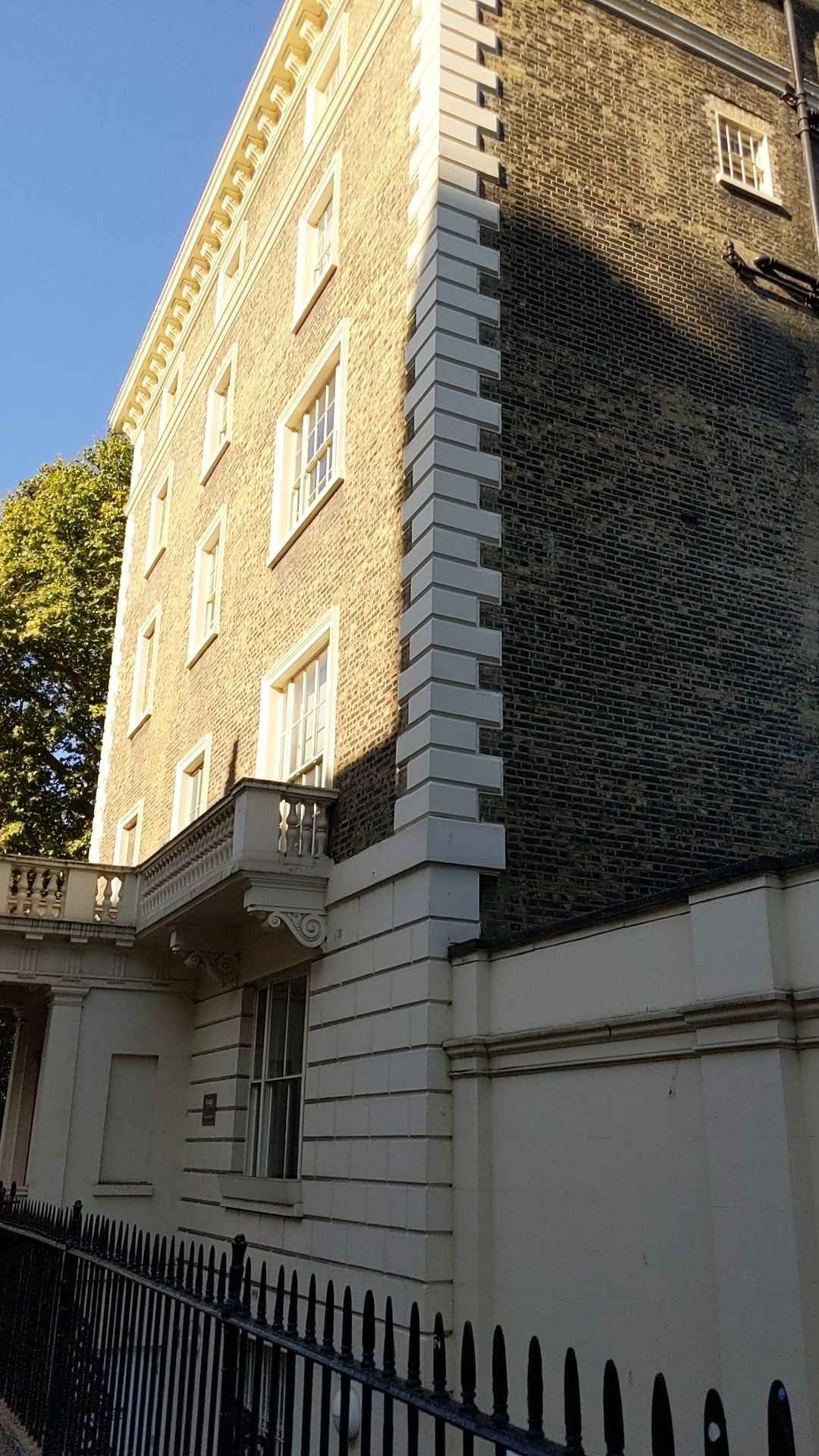}
\end{minipage}
\begin{minipage}{0.32\textwidth}\centering
    \includegraphics[width=0.8\linewidth]{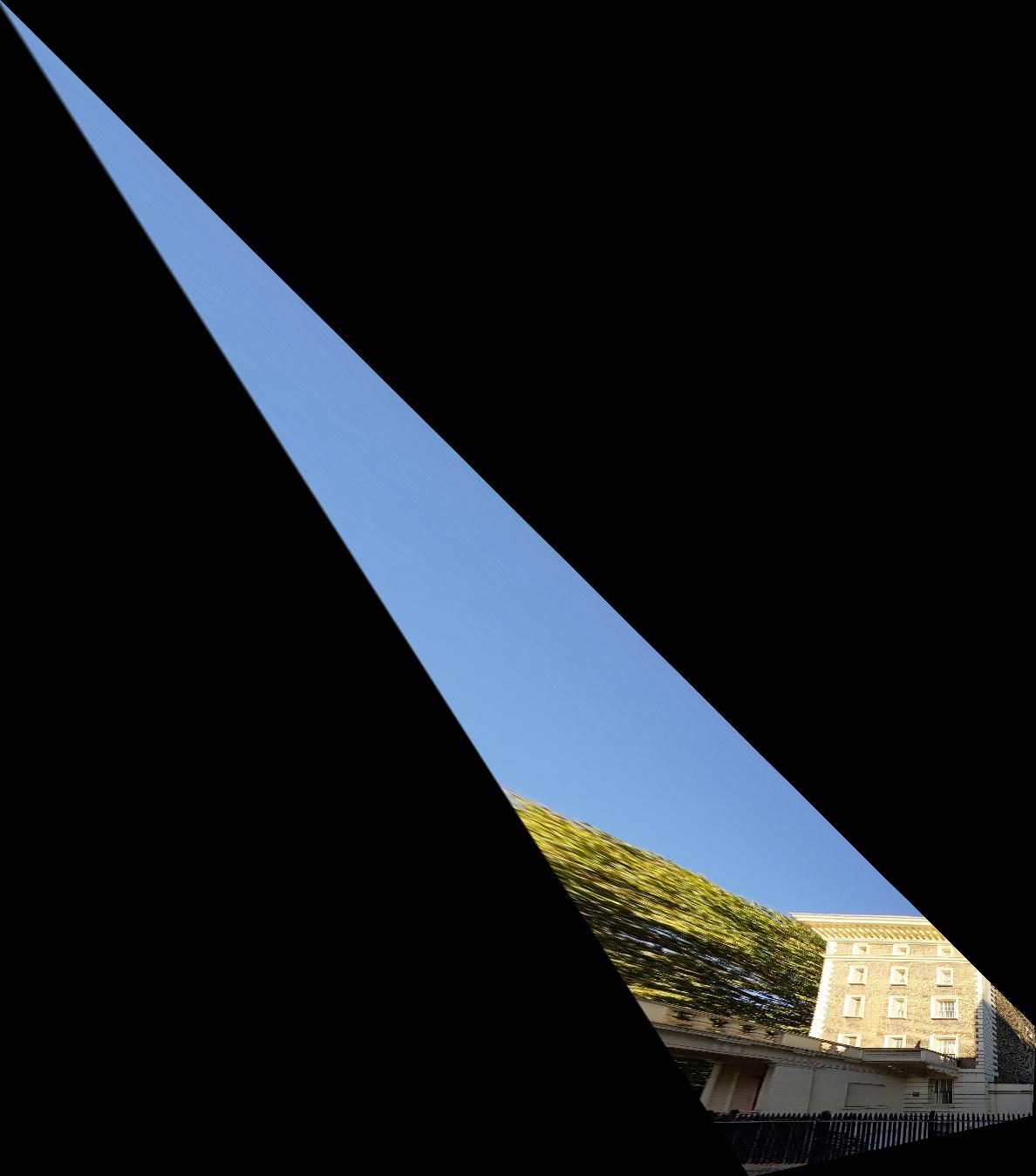}
\end{minipage}
\begin{minipage}{0.32\textwidth}\centering
    \includegraphics[width=0.8\linewidth]{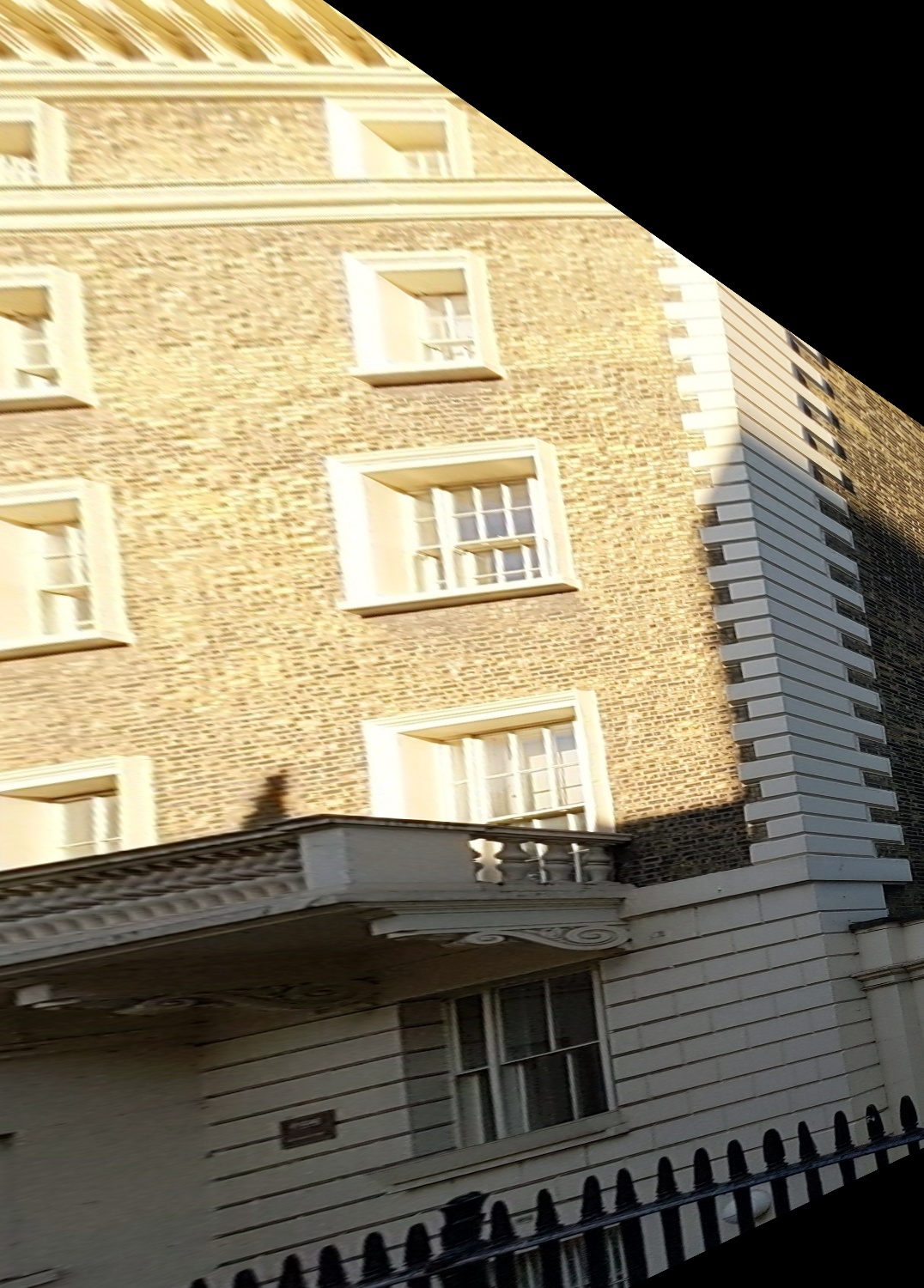}
\end{minipage}
\\
\begin{minipage}{0.32\textwidth}\centering
    \footnotesize{
    Input image
    \\
    (1920 x 1080)
    }
\end{minipage}
\begin{minipage}{0.32\textwidth}\centering
    \footnotesize{
    Rectified with VP
    \\
    (1349 x 1188)
    }
\end{minipage}
\begin{minipage}{0.32\textwidth}\centering
    \footnotesize{
    Rectified with Ours
    \\
    (1500 x 1076)}
\end{minipage}

    \caption{Perspective rectification of a challenging example. From left to right: input image, perspective rectification with a vanishing point method~\cite{chaudhury2014auto} and Ours. The vanishing point method can heavily distort an image and produce an image where the region of interest only occupies a small part. An output image may become prohibitively large in order to preserve detail. Our method does not have these artifacts as it rectifies planar patches, not the full image. }
    \label{fig:vp-rect-distortion}
\end{figure}

Even without known 3D scene geometry, it is still possible to remove the effect of perspective distortion from a single image~\cite{Robertson2004BMVC,Baatz2010ECCV,Pritts2018ACCV}.  
In principle, vanishing points~\cite{Chaudhury2014ICIP,Li2019ICCV,Zhou2019NeurIPS,Simon2018ECCV} or repeating structural elements~\cite{Pritts2018ACCV,Pritts2018CVPR,Wu2010ECCV} can be used to rectify planar regions prior to feature detection~\cite{Criminisi2000IJCV}. 
However, this process is cumbersome in practice: 
it is unclear which pixels belong to a plane, so it is necessary to unwarp the full image. 
This introduces strong distortions for image regions belonging to different planes. 
As a result, determining a good resolution for the unwarped image is a challenge, since one would like to avoid both too small (resulting in a loss of details) and too high resolutions (which quickly become hard to handle). Fig. \ref{fig:vp-rect-distortion} shows an example of this behaviour on an image from our dataset. 
This process has to be repeated multiple times to handle multiple planes. 

Prior work has shown the advantages of removing perspective distortion prior to feature detection in tasks such as visual localization~\cite{Robertson2004BMVC} and image retrieval~\cite{Baatz2010ECCV,Chen2011CVPR,Baatz2012IJCV}. 
Yet, such methods are not typically used in practice as they are hard to automate. 
For example, modern SfM~\cite{schoenberger2016sfm,Wu20133DV},  SLAM~\cite{MurArtal2017TRO}, and visual localization~\cite{Sattler2017PAMI,Sarlin2019CVPR,Germain20193DV} systems still rely on classical features without any prior removal of perspective effects. 
This paper shows that convolutional neural networks (CNNs) for single-image depth estimation provide a simple yet effective solution to the practical problems encountered when correcting perspective distortion: 
although their depth estimates might be noisy (especially for scene geometry far away from the camera), they are typically trained to produce smooth depth gradients~\cite{megadepth,monodepth1}. 
This fact can be used to estimate normals, which in turn define planes that can be rectified. 
Per-pixel normals provide information about which pixels belong to the same plane, thus avoiding the problems of having to unwarp the full image and to repeatedly search for additional planes. 
Also, depth information can be used to avoid strong distortions by ignoring pixels seen under sharp angles. 
As  a result, our approach is significantly easier to use in practice. 

Specifically, this paper makes the following contributions: 
\textbf{1)} We propose a simple and effective 
method for removing perspective distortion prior to feature extraction based on single-image depth estimation. 
\textbf{2)} We demonstrate the benefits of this approach through detailed experiments on feature matching and visual localization under strong viewpoint changes. 
In particular, we show that improved performance does not require fine-tuning the depth prediction network per scene. 
\textbf{3)} We propose a new dataset to evaluate the performance of feature matching under varying viewpoints and viewing conditions. 
We show that our proposed approach can significantly improve matching performance 
in the presence of dominant planes, without significant degradation if there are no dominant planes. 

\section{Related Work}

\PAR{Perspective undistortion via 3D geometry.} 
If the 3D geometry of the scene is known, \eg, from depth maps recorded by RGB-D sensors, laser scans, or multi-view stereo, it is possible to remove perspective distortion prior to feature extraction~\cite{wu20083d,Koeser2007ICCV,Zeisl2012ECCVW,Zeisl2013ICCV,Wu2008CVPRW}. More precisely, each plane detected in 3D defines homographies that warp perspective image observations of the plane to orthographic projections. 
Extracting features from orthographic rather than perspective views makes the features (theoretically) invariant to viewpoint changes\footnote{In practice, strong viewpoint changes create strong distortions in the unwarped images, which prevent successful feature matching~\cite{Koeser2007ICCV}.}.
This enables feature matching under strong viewpoint changes~\cite{Zeisl2013ICCV}. 
Given known 3D geometry, a single feature match between two images or an image and a 3D model can be sufficient to estimate the full 6-degree-of-freedom (relative) camera pose~\cite{wu20083d,Baatz2010ECCV}. 
Following similar ideas, features found on developable surfaces can be made more robust by unrolling the surfaces into a plane prior to feature detection. 
Known 3D geometry can also be used to remove the need for certain types of invariances~\cite{Koeser2007ICCV},
\eg, predicting the scale of keypoints from depth~\cite{Jones2011IJRR} removes the need for scale invariance.

Previous work assumed that 3D data is provided together with an image, or is extracted from multiple images. 
Inspired by this idea, we show that perspective distortion can often be removed effectively using single-image depth predictions made by modern convolutional neural networks (CNNs).

\PAR{Perspective undistortion without 3D geometry.} 
Known 3D geometry is not strictly necessary to remove the effect of perspective foreshortening on planar structures: 
either vanishing points~\cite{Chaudhury2014ICIP,Li2019ICCV,Zhou2019NeurIPS,Simon2018ECCV,Criminisi2002DAGM} or repeating geometrical structures~\cite{Pritts2018ACCV,Pritts2018CVPR,Pritts2016BMVC,Pritts2014CVPR,Wu2010ECCV} can be used to define a homography for removing perspective distortion for all pixels on the plane~\cite{Criminisi2000IJCV,Liebowitz1999CGF}. 
In both cases, an orthographic view of the plane can be recovered up to an unknown scale factor and an unknown in-plane rotation, \ie, an unknown similarity transformation. 
Such approaches have been used to show improved performance for tasks such as image retrieval \cite{Baatz2010ECCV,Baatz2012IJCV,Chen2011CVPR},  visual localization~\cite{Robertson2004BMVC}, and feature matching~\cite{Cao2009WACV}. 
However, they are often brittle and hard to automate for practical use: 
they do not provide any information about which pixels belong to a given plane. 
This makes it necessary to warp the full image, which can introduce strong distortion effects for regions that do not belong to the plane. 
This in turn leads to the problem of selecting a suitable resolution for the unwarped image, to avoid loosing details without creating oversized images that cannot be processed efficiently. 
As a result, despite their expected benefits, such methods have seen little use in practical applications, \eg, modern SfM or SLAM systems. See Fig. \ref{fig:vp-rect-distortion} (as well as Sec. 8 in the supplementary material) for an example of this behaviour as seen on an image from our dataset.

In this paper, we show that these problems can easily be avoided by using single-image depth predictions to remove the effect of perspective distortion. 

\PAR{Pairwise image matching via view synthesis.} 
An alternative to perspective undistortion for robust feature matching between two images taken from different viewpoints is view synthesis~\cite{morel2009asift,mishkin2015mods,Pang2012Neurocomputing,Liu2012MVA}. 
Such approaches generate multiple affine or projective warps of each of the two images and extract and match features for each warp. 
Progressive schemes exist, which first evaluate small warps and efficient features to accelerate the process~\cite{mishkin2015mods}. 
Still, such approaches are computationally very expensive due to the need to evaluate a large number of potential warps. 
Methods like ours, based on removing perspective distortion, avoid this computational cost by determining a single warp per region. 
Such warps can also be estimated locally per region or patch~\cite{Altwaijry2016CVPR,Hinterstoisser2011IJCV}. 
This latter type of approach presupposes that stable keypoints can be detected in perspectively distorted images. 
Yet, removing perspective effects prior to feature detection can significantly improve performance~\cite{Wu2008CVPRW,Koeser2007ICCV}. 





\PAR{Datasets.} 
Measuring the performance of local features under strong viewpoint changes, \ie, the scenario where removing perspective distortion could provide the greatest benefit, has a long tradition, so multiple datasets exist for this task~\cite{Mikolajczyk2005IJCV,balntas2017hpatches,Shao2003TR,mishkin2015mods,cordes2013high,aanaes2012interesting,Altwaijry2016CVPR}. 
Often, such datasets depict nicely textured scenes, \eg, graffiti, paintings, or photographs, from different viewpoints. 
Such scenes represent ``failure cases'' for single-image depth predictions as the networks (not unreasonably) predict the depth of the elements shown in the graffiti \etc (\cf Fig.~\ref{fig:hpatches:examples}). 
This paper thus also contributes a new dataset for measuring performance under strong viewpoint changes that depicts regular street scenes. 
In contrast to previous datasets, \eg, \cite{balntas2017hpatches}, ours contains both viewpoint and appearance changes occuring at the same time.


\PAR{Single-image depth prediction.}
Monocular depth estimation aims at training a neural network to predict a depth map from a single RGB image. 
Supervised methods directly regress ground-truth depth,  acquired with active sensors (LiDAR or Kinect)~\cite{eigen2014depth,planenet,planercnn}; from SfM reconstructions~\cite{megadepth}; or manual ordinal annotations~\cite{depthinthewild, megadepth}. 
However, collecting training data is difficult, costly, and time-consuming. 
Self-supervised training minimizes a photometric reprojection error between views. These views are either frames of videos~\cite{monodepth2, klodt}, and/or stereo pairs~\cite{garg2016unsupervised, monodepth1, monodepth2, depth-hints}. Video-only training also needs to estimate the pose between frames (up to scale) and model moving objects~\cite{zhou2017CVPR}. Training with stereo provides metric-accurate depth predictions if the same camera is used at test time. 
Supervised and self-supervised losses can be combined during training~\cite{kuznietsov, dvso}. 


CNNs can be trained to predict normals~\cite{eigen2015predicting,wang2015designing,hickson2019floors,dhamo2019object}, both depth and normals~\cite{lego, yin2019enforcing, li2015depth, zhan2019self}, or 3D plane equations~\cite{planenet, planercnn}. However, normals are either used to regularize depth, or trained exclusively on indoor scenes because of availability of supervised data, which is difficult to collect for outdoor scenes~\cite{chen2017surface}. 

The approach presented in this paper is not tied to any specific single view depth prediction approach, and simply assumes that approximate depth information is available. Normal estimation networks could also be used in our pipeline, however depth estimation networks are more readily available.

\section{Perspective Unwarping}
\label{sec:method}
We introduce a method for performing perspective correction of monocular images. The aim is to perform this prior to feature extraction, leading to detection and description of features that are more stable under viewpoint changes. That stability, can for example, establish more numerous correct correspondences between images taken from significantly different viewpoints.

The method is inspired by, and bears close resemblance to, the view-invariant patch descriptor by 
Wu~\etal~\cite{wu20083d}. The main difference is that while Wu's method was designed for alignment of 3D point clouds, our method can be applied to single, monocular images, allowing it to be utilized in applications such as  wide-baseline feature matching, single image visual localization, or structure from motion. 

\begin{figure}[t]
\begin{minipage}{\textwidth}
    \centering
    \includegraphics[width=\linewidth]{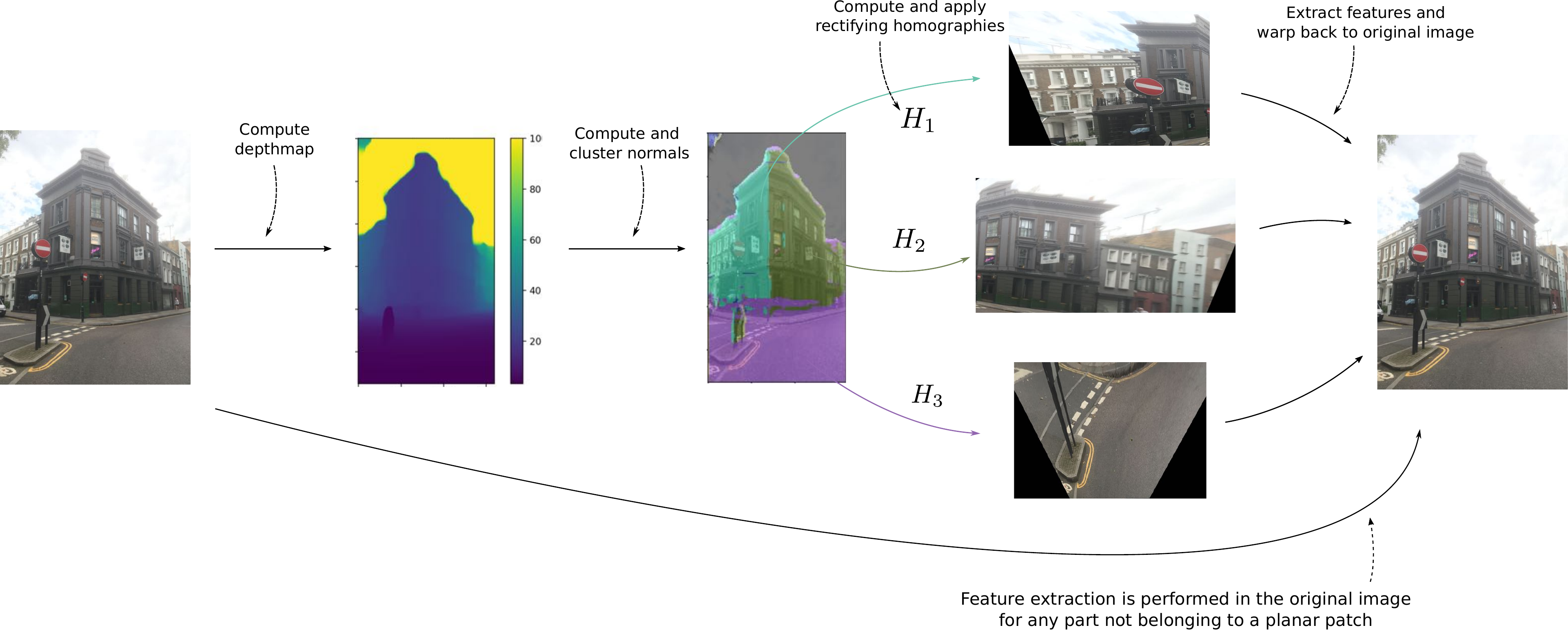}
    \caption{Proposed pipeline for extracting perspectively corrected features from a single image: Depths are computed using a single-image depth estimation network. The intrinsic parameters of the camera are then used to backproject each pixel into 3D space, and a surface normal is estimated for each pixel. The normals are clustered into three orthogonal directions, and a homography is computed for each cluster to correct for the perspective distortion. Local image features are then extracted from each rectified patch using an off-the-shelf feature extractor and their positions are warped back into the original image. Regular local features are extracted from the parts of the image that do not belong to a planar patch. } 
    \label{fig:method:overview}
\end{minipage}
\end{figure}

A schematic overview of the method is shown in Fig. \ref{fig:method:overview}. The central idea is that given a single input image, a network trained for single image depth estimation is used to compute a corresponding dense depth map of the image. Using the camera intrinsics, the depth map gets backprojected into a point cloud, given in the camera's reference frame. From this point cloud, a surface normal vector is estimated for each point. For any given point in the point cloud (and correspondingly, in the image), a rectifying homography $\mathit{H}$ can now be computed. $\mathit{H}$ transforms a patch centered around the point in the image to a corresponding patch, which simulates a virtual camera looking straight down on the patch, \ie a camera whose optical axis coincides with the patch's surface normal. As shown by several experiments in Sec.~\ref{sec:experiments}, by performing both feature detection and description in this rectified space, the obtained interest points can sometimes be considerably more robust to viewpoint differences.

In principle this method could be applied to each point independently, but typically a large number of points will share the same normal. Consider for example points lying on a plane, such as points on the ground, points on the same wall, or points on different but parallel planes, such as opposing walls. These points will all be rectified by the same homography. This is utilized in the proposed method by identifying planar regions in the input image, which is done by clustering all normals on the unit sphere. This yields a partitioning of the input image into several connected components, which are then rectified individually. Each input image is thus transformed into a set of rectified patches, consisting of perpendicular views of all 
dominant planes in the image. Image features can then be extracted, using any off-the-shelf detector and descriptor. In the experiments, results are provided for SIFT \cite{lowe2004distinctive}, SuperPoint~\cite{detone2018superpoint}, ORB~\cite{rublee2011orb}, and 
BRISK\cite{leutenegger2011brisk} features.

Note that the rectification process is not dependent on all planes being observed. If only one plane is visible, that may still be rectified on its own. Parts of the image that are not detected to be on a plane are not rectified, but we still extract features from these parts and use them for feature matching. This way, we do not ignore good features just because they are not on a planar surface. For complex, non-planar geometries, large parts of an image may not have planar surfaces. For such images our approach gracefully resorts to standard (non-rectified) feature matching for regions not belonging to the identified planes.

Below, we describe each of the above steps in more detail. 

\subsection{Depth Estimation}
The first step in the perspective correction process is the computation of the depth map. In this paper we use MonoDepth2~\cite{monodepth2} which was trained with a Depth Hints loss~\cite{depth-hints} on several hours of stereo video captured in one European city and three US cities. In addition to stereo, the network was also trained on the MegaDepth dataset~\cite{megadepth} and Matterport~\cite{matterport} datasets (see supplementary materials for details). 
This network takes as input a single image resized to $512 \times 256$, and outputs a dense depth map. Under ideal conditions, each pixel in the depth map tells us the calibrated depth in meters. In practice, any method that provides dense depth estimates may be used, and the depths need not be calibrated, \ie depths estimated up to an unknown scale factor may also be used, since the depth map is only used to compute surface normals for each point. 

\subsection{Normal Computation and Clustering}
With the depth map computed, the next step is normal computation. A surface normal is estimated for each pixel in the depth map by considering a $5 \times 5$ window centered on the pixel, and fitting a plane to the 25 corresponding back-projected points. The unit normal vector of the plane is taken as an estimate of the surface normal for that pixel. 

With the normals computed, they are then clustered to identify regions in the image corresponding to planar surfaces. Since all points on the same plane share the same normals, these normals (and the pixels assigned to them) may be found by performing $k$-means clustering. 

Since the depth map, and by extension the surface normals, are subject to noise, we found that clustering the normals into three clusters or dominant directions, while also enforcing orthogonality between these clusters, gave good results. 
Each cluster also includes its antipodal point (this means, for example, that two opposing walls would be assigned to the same dominant direction). This assumption seems to correspond to the 3D structure of many scenes: if at least one dominant planes is visible, such as the ground or a building wall, this method will produce satisfactory results. If two are visible, the estimated normals, and thus also the estimated homographies, tend to be more accurate. If no planes are visible, the method gracefully reduces to regular feature extraction. Note also that several different patches in the image can be assigned to the same cluster, but rectified separately as different planes: examples include opposing walls or parallel flat surfaces. 



In non-Manhattan world geometries, where several non-perpendicular planes are visible, the estimated normals may not be completely accurate. Thus, our method would apply a homography that would render the planar surface not from a fronto-parallel view, but at a tilt. In most cases, this rectification still removes some effects of perspective distortion. 

\subsection{Patch Rectification}
With the normals clustered into three dominant clusters, 
each pixel is assigned its normal's respective cluster. Each of these subsets may be further subdivided into their respective connected components. The input image is thus partitioned into a set of patches, each consisting of a connected region of pixels in the image, together with a corresponding estimate of the surface normal for that patch. In Fig.~\ref{fig:method:overview}, the patches are shown overlaid on the image in different colors. 

A rectified view of each patch is now computed, using the estimated patch normal. The patch is warped using a homography, computed as the homography which maps the patch to the patch as it would have been seen in a virtual camera sharing the same camera center as the original camera, but rotated such that its optical axis is parallel to the surface normal (\ie it is facing the patch straight on). The smallest rotation that brings the camera into this position is used. 

Lastly, not the entire patch is rectified, since if the plane corresponding to a given patch is seen at a glancing angle in the camera, most of the rectified patch would be occupied by heavily distorted, or stretched, regions. As such, a threshold of $80^{\circ}$ is imposed on the maximum angle allowed between the viewing ray from the camera, and the surface normal, and the resulting patch is cropped to only fully contain the region of the patch seen at not too glancing an angle.

\subsection{Warping Back}
When matching features, the image may now be replaced with its set of rectified patches and patches from non-planar parts of the image. Alternatively, feature extraction may be performed in the non-planar parts of the original image, and in all rectified patches, and the 2D locations of the features in the rectified patches may then be warped back into the original image coordinate system, but with the descriptors unchanged. A perspectively corrected representation of the image has then been computed. The final description thus includes perspectively corrected features for all parts of the image that were deemed as belonging to a plane, and regular features extracted from the original image, from the parts that were deemed non-planar.

\section{Dataset for Strong Viewpoint Changes}
\label{sec:dataset}
\begin{figure}[t]
    \centering
    \begin{minipage}{\linewidth}
    \includegraphics[width=0.16\linewidth]{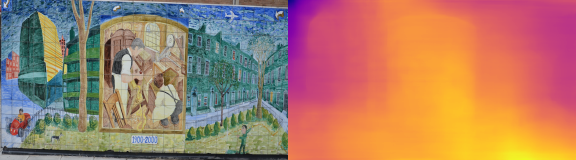}
    \includegraphics[width=0.16\linewidth]{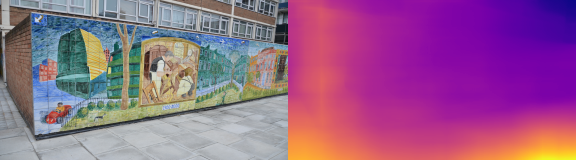}
    \includegraphics[width=0.16\linewidth]{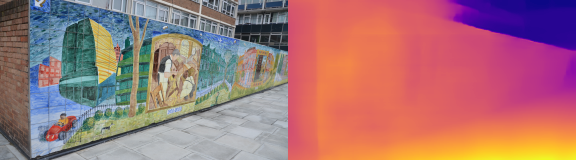}
    \includegraphics[width=0.16\linewidth]{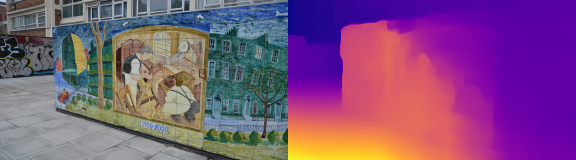}
    \includegraphics[width=0.16\linewidth]{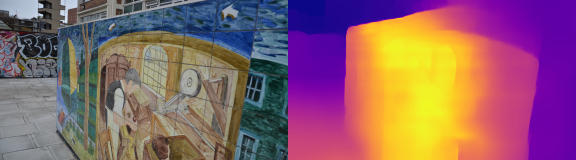}
    \includegraphics[width=0.16\linewidth]{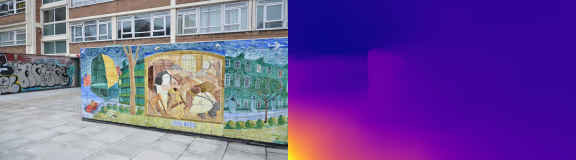}
    \end{minipage}
    \\
    \includegraphics[width=0.16\linewidth]{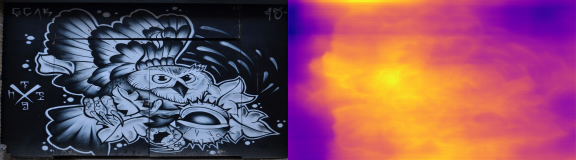}
    \includegraphics[width=0.16\linewidth]{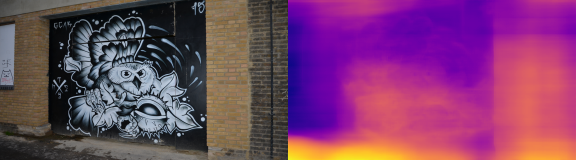}
    \includegraphics[width=0.16\linewidth]{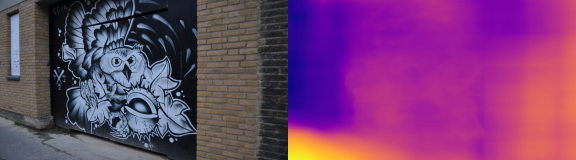}
    \includegraphics[width=0.16\linewidth]{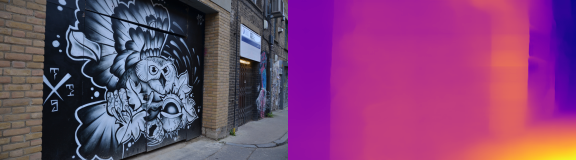}
    \includegraphics[width=0.16\linewidth]{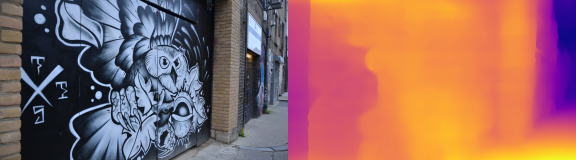}
    \includegraphics[width=0.16\linewidth]{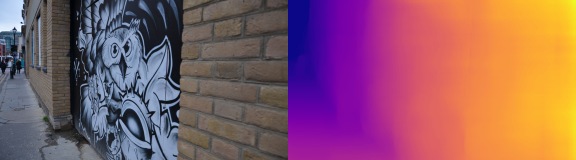}
    \\
    \includegraphics[width=0.16\linewidth]{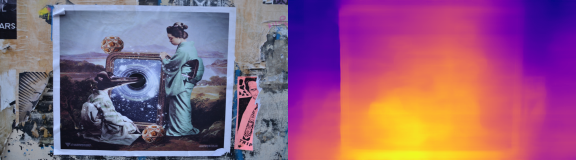}
    \includegraphics[width=0.16\linewidth]{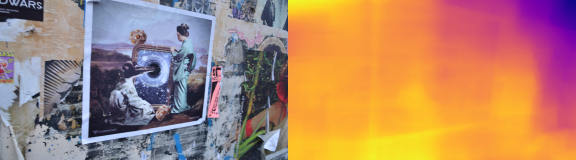}
    \includegraphics[width=0.16\linewidth]{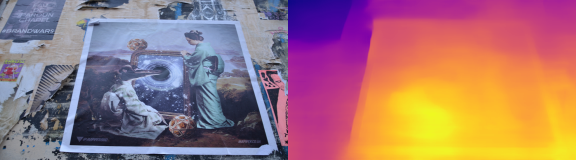}
    \includegraphics[width=0.16\linewidth]{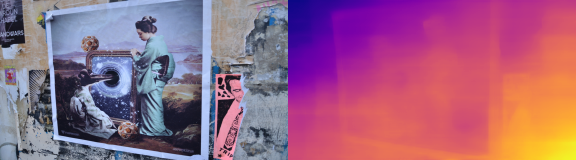}
    \includegraphics[width=0.16\linewidth]{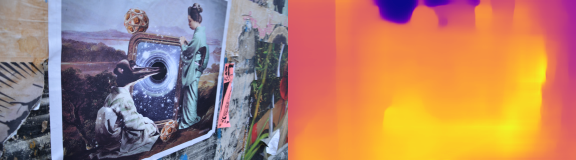}
    \includegraphics[width=0.16\linewidth]{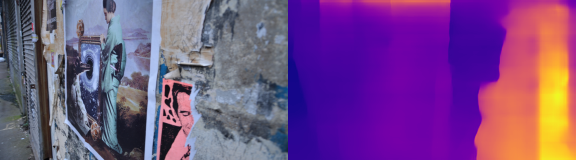}
    \\
    \includegraphics[width=0.16\linewidth]{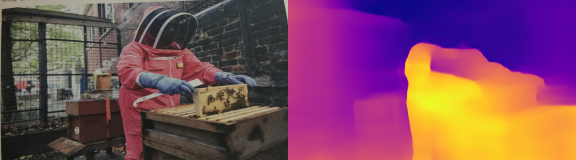}
    \includegraphics[width=0.16\linewidth]{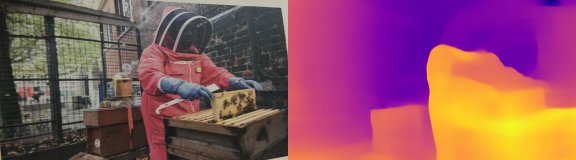}
    \includegraphics[width=0.16\linewidth]{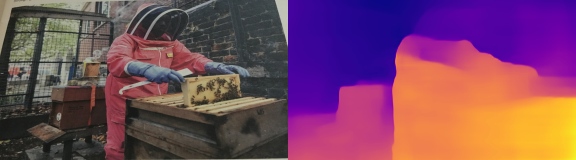}
    \includegraphics[width=0.16\linewidth]{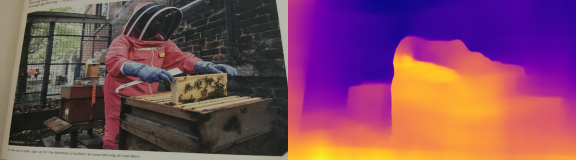}
    \includegraphics[width=0.16\linewidth]{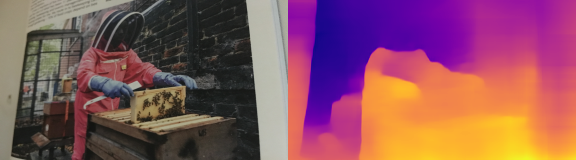}
    \includegraphics[width=0.16\linewidth]{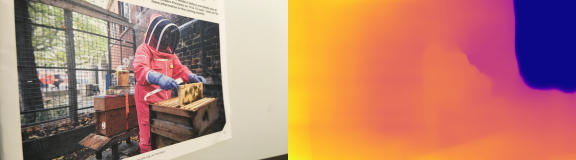}
    \caption{``Failure'' cases of applying a modern single-image depth prediction network~\cite{monodepth2,depth-hints} on images from the HPatches dataset~\cite{balntas2017hpatches} (visualized as inverse depth): the network predicts the depth of the scenes depicted in the graffiti and images rather than understanding that these are drawings / photos attached to a planar surface.}
    \label{fig:hpatches:examples}
\end{figure}

A modern, and already well-established, benchmark for evaluating local features is the HPatches dataset~\cite{balntas2017hpatches}. 
HPatches consists of 116 sequences of 6 images each, where sequences have either illumination or viewpoint changes. 
Similar to other datasets such as~\cite{mikolajczyk2005performance}, planar scenes that can be modeled as homographies are used for viewpoint changes. 
Most of the sequences depict paintings or drawings on flat surfaces. 
Such scenes are ideal for local features as they provide abundant texture. 
Interestingly, such scenes cause single-image depth prediction to ``fail'': 
as shown in Fig.~\ref{fig:hpatches:examples}, networks predict the depth of the structures shown in the paintings and drawings rather than modeling the fact that the scene is planar. 
We would argue that this is a rather sensible behavior as the scene's planarity can only be inferred from context, \eg, by observing a larger part of a scene rather than just individual drawings. 
Still, this behavior implies that standard datasets are not suitable for evaluating the performance of any type of method based on singe-image depth prediction. This motivated us to capture our own dataset that, in contrast to 
benchmarks such as HPatches, intentionally contains non-planar structures.

In this paper, we thus present a new dataset for evaluating the robustness of local features when matching across large viewpoint variations, and changes in lighting, weather, \etc. 
The dataset consists of 8 separate scenes, where each scene consists of images of one facade or building captured from a wide range of different viewpoints, and in different weather and environmental conditions. Each scene has been revisited up to 5 times, see Tab.~\ref{tab:my_label}. 

\begin{figure}[!ht]
\begin{minipage}{\linewidth}
    \centering
    \begin{minipage}{\linewidth}
    \includegraphics[width=0.115\linewidth]{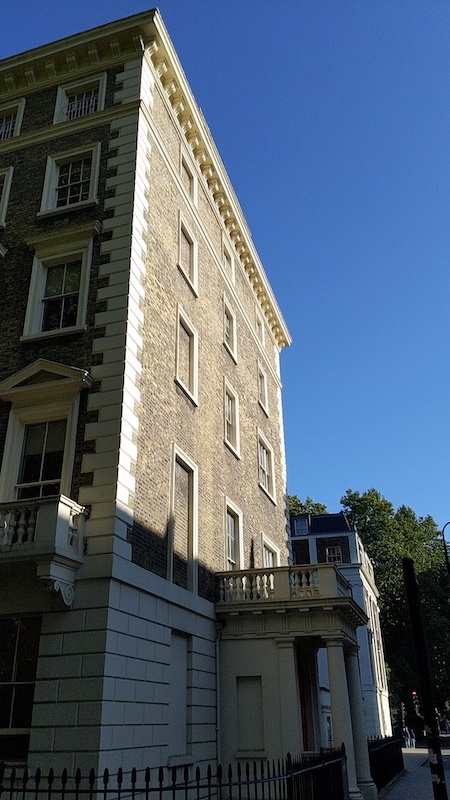}
    \includegraphics[width=0.115\linewidth]{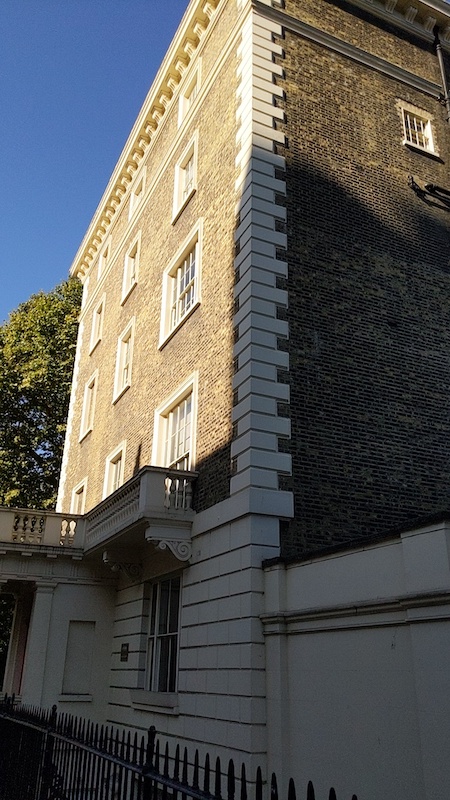}
    \includegraphics[width=0.115\linewidth]{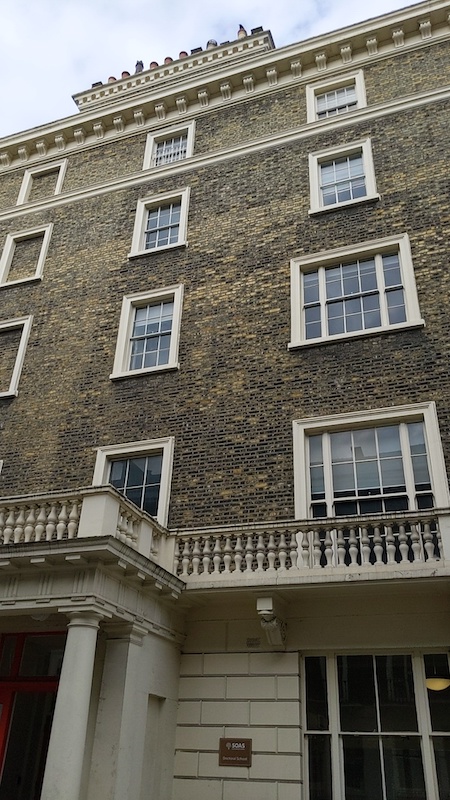}
    \includegraphics[width=0.115\linewidth]{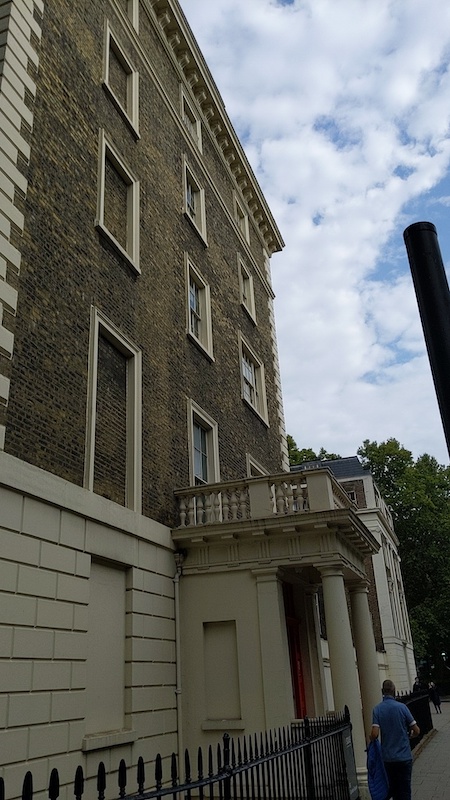}
    \hspace{1pt}
    \includegraphics[width=0.23\linewidth]{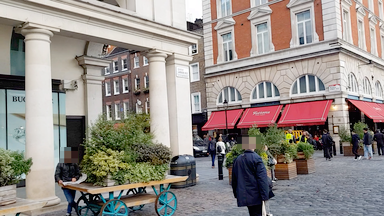}
    \includegraphics[width=0.23\linewidth]{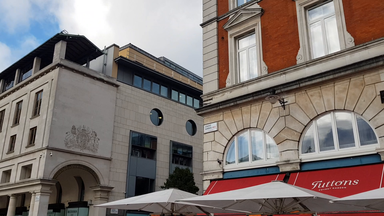}
    \end{minipage}
    \\
\end{minipage}
    \caption{ 
    Six example images, showing two different scenes of the presented benchmark dataset. The dataset contains several scenes, each consisting of over 1000 images of an urban environment captured during different weather conditions, and covering a large range of viewing angles. This dataset permits the evaluation of the degradation of local feature matching methods for increasing viewpoint angle differences. Please see the supplementary material for example images from each of the eight scenes. }
    \label{fig:dataset:examples}
\end{figure}

All images included in the dataset originated from continuous video sequences captured using a consumer smartphone camera. These video sequences were then reconstructed using the Colmap software \cite{schoenberger2016sfm, schoenberger2016mvs}, and the poses for a subset of the images were extracted from this reconstruction. 
Colmap also provides an estimate of the intrinsic parameters for each individual image frame, which are included in the dataset. 
These are necessary since the focal length of the camera may differ between the images due to the camera's autofocus. 
Fig.~\ref{fig:dataset:examples} shows a set of example images from two of our 8 scenes.

Since our scenes are not perfectly planar, measuring feature matching performance by the percentage of matches that are inliers to a homography, as done in~\cite{balntas2017hpatches,mikolajczyk2005performance}, is not an option for our dataset. 
Inspired by 
CVPR 2019 workshops ``Image Matching: Local Features and Beyond'' and ``Long-Term Visual Localization under Changing Conditions'', we evaluate feature matching on downstream tasks as opposed to measuring the number of recovered feature matches or repeatability, \etc.
Hence, we evaluate the performance of local features through the task of accurately estimating the relative pose between pairs of images. This allows us to judge if improvements in feature matching lead to  meaningful improvements in practical applications such as localization and structure from motion. 


For each scene, a list of image pairs is thus provided. 
Each image pair has been assigned to one of eighteen different difficulty categories, depending on the distance between the centres of the cameras that captured the images, and the magnitude of their relative rotation. The difficulty categories span the range of almost no difference in rotation up to almost $180^{\circ}$ relative rotation. So, the image pairs in the $k$-th difficulty category have a relative rotation in the range of $[10k, 10(k+1)]$ degrees in one of the axes. 

\begin{table}[t]
    \centering
    \caption{Statistics for the scenes in our dataset}
    \scriptsize{
    \setlength{\tabcolsep}{3pt}
    \begin{tabular}{| l | c | c | c | c | c | c | c | c |}
        \hline
        scene \# & 1 & 2 & 3 & 4 & 5 & 6 & 7 & 8 \\ \hline 
        \# img. pairs & 3590 & 3600 & 3600 & 2612 & 3428 & 3031 & 2893 & 3312 \\ \hline
        \# sequences & 4 & 4 & 3 & 4 & 3 & 5 & 3 & 3 \\ \hline
    \end{tabular}
    }%
    \label{tab:my_label}
\end{table}





The dataset is publicly available on the project webpage \url{www.github.com/nianticlabs/rectified-features}.

\section{Experiments} 
\label{sec:experiments}
This section provides two experiments: 
Sec.~\ref{sec:experiments:matching} shows that perspective unwarping based on the proposed approach can significantly improve feature matching performance on our proposed dataset. 
Sec.~\ref{sec:experiments:localization} shows that our approach can be used to re-localize a car under 180$^\circ$ viewpoint changes, \eg, in the context of loop closure handling for SLAM. 
%
%
We use the SIFT~\cite{lowe2004distinctive} implementation provided by OpenCV~\cite{opencv_library} for all of our experiments.

\subsection{Matching Across Large Viewpoint Changes}
\label{sec:experiments:matching}
First, we evaluate our method on the 8 scenes of our proposed dataset. As a baseline we evaluate the performance of traditional and recently proposed learned local image features. To demonstrate the benefit of perspective rectification, we perform image matching with the same set of local features on the same set of image pairs. 

For all image pairs in the dataset, feature matching was performed using SIFT\cite{lowe2004distinctive}, SuperPoint\cite{detone2018superpoint}, ORB~\cite{rublee2011orb} and BRISK\cite{leutenegger2011brisk} features. For each feature type, feature matching was performed between features extracted from the original images, as well as between the perspectively corrected features, as explained in Sec.~\ref{sec:method}. Our unoptimized implementation performs image rectification in around 0.8 seconds per image. Using the established matches and the known intrinsics of the images, an essential matrix was computed, and the relative camera pose was then retrieved from this. This relative pose was compared to the ground truth relative pose (computed by Colmap as described in Sec. \ref{sec:dataset}). An image pair was considered successfully localized if the difference between the estimated relative rotation and the ground truth relative rotation was smaller than $5^{\circ}$, where the difference between two rotations is taken as the magnitude of the smallest rotation that aligns one rotation with the other. Also included is a curve showing the performance of SIFT features extracted from images rectified using a vanishing point-based rectification method \cite{chaudhury2014auto}. 

\begin{figure}[t]
    \centering
    \includegraphics[width=0.32\linewidth]{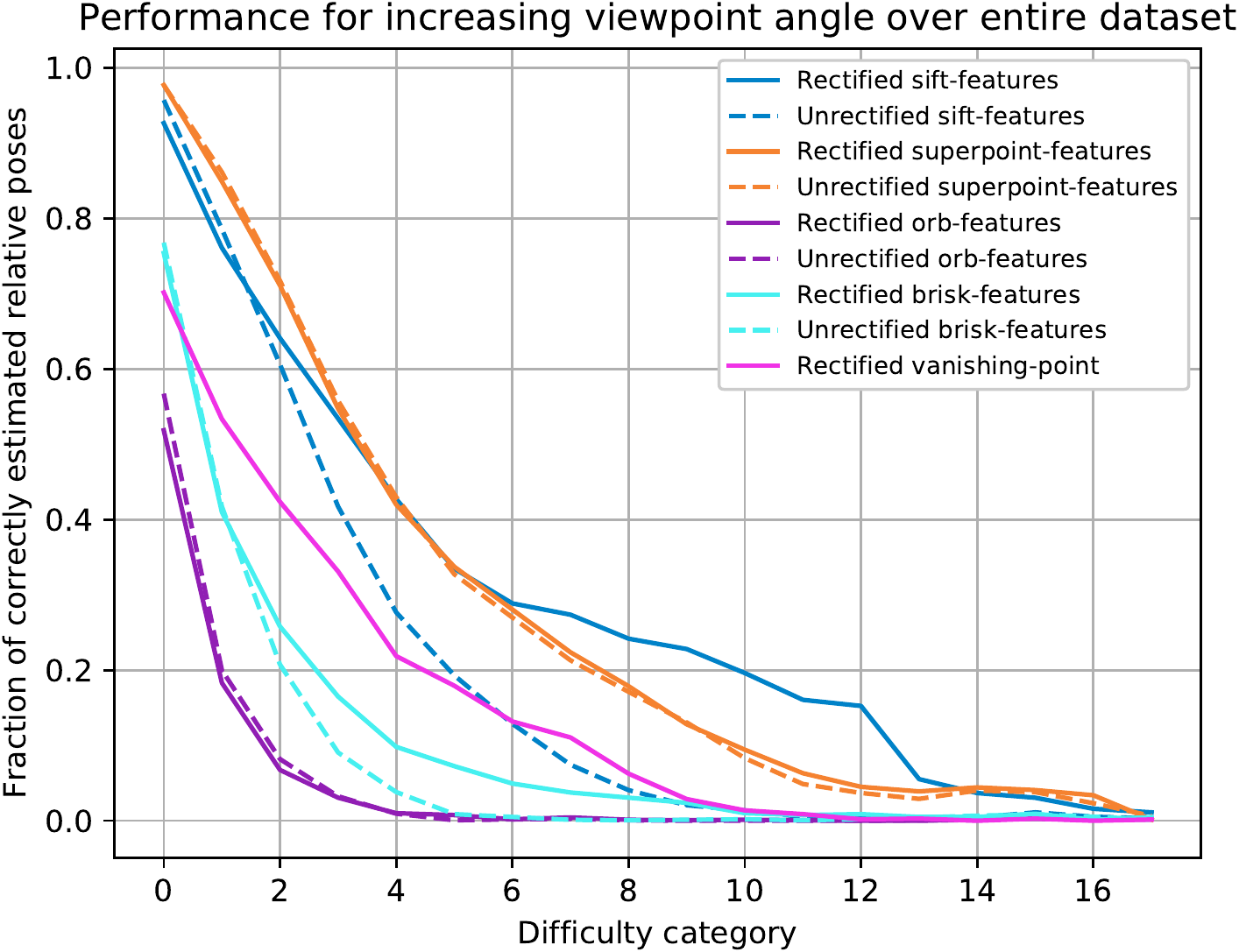}\hspace{0.5em}
    \includegraphics[width=0.32\linewidth]{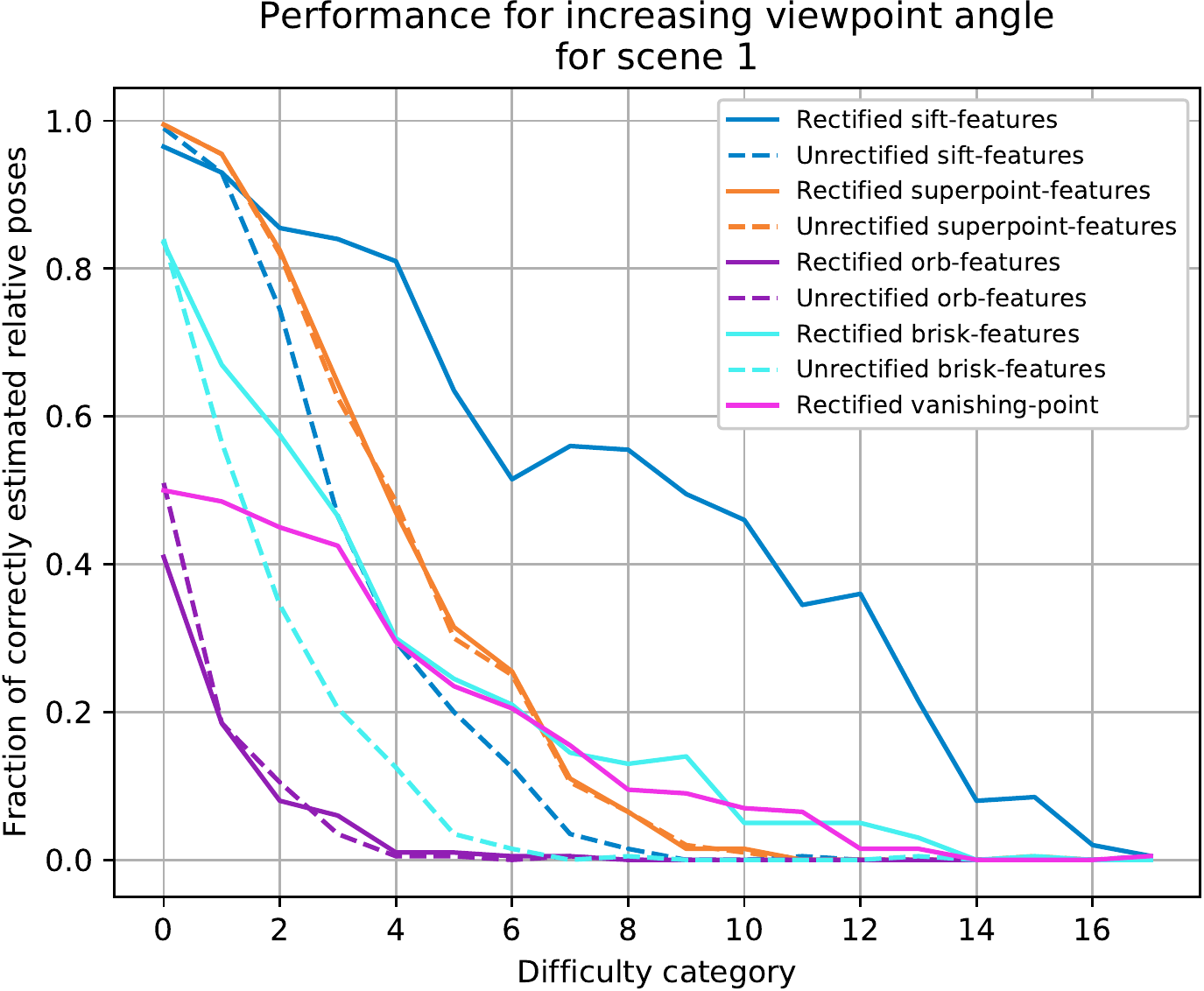}
    \includegraphics[width=0.32\linewidth]{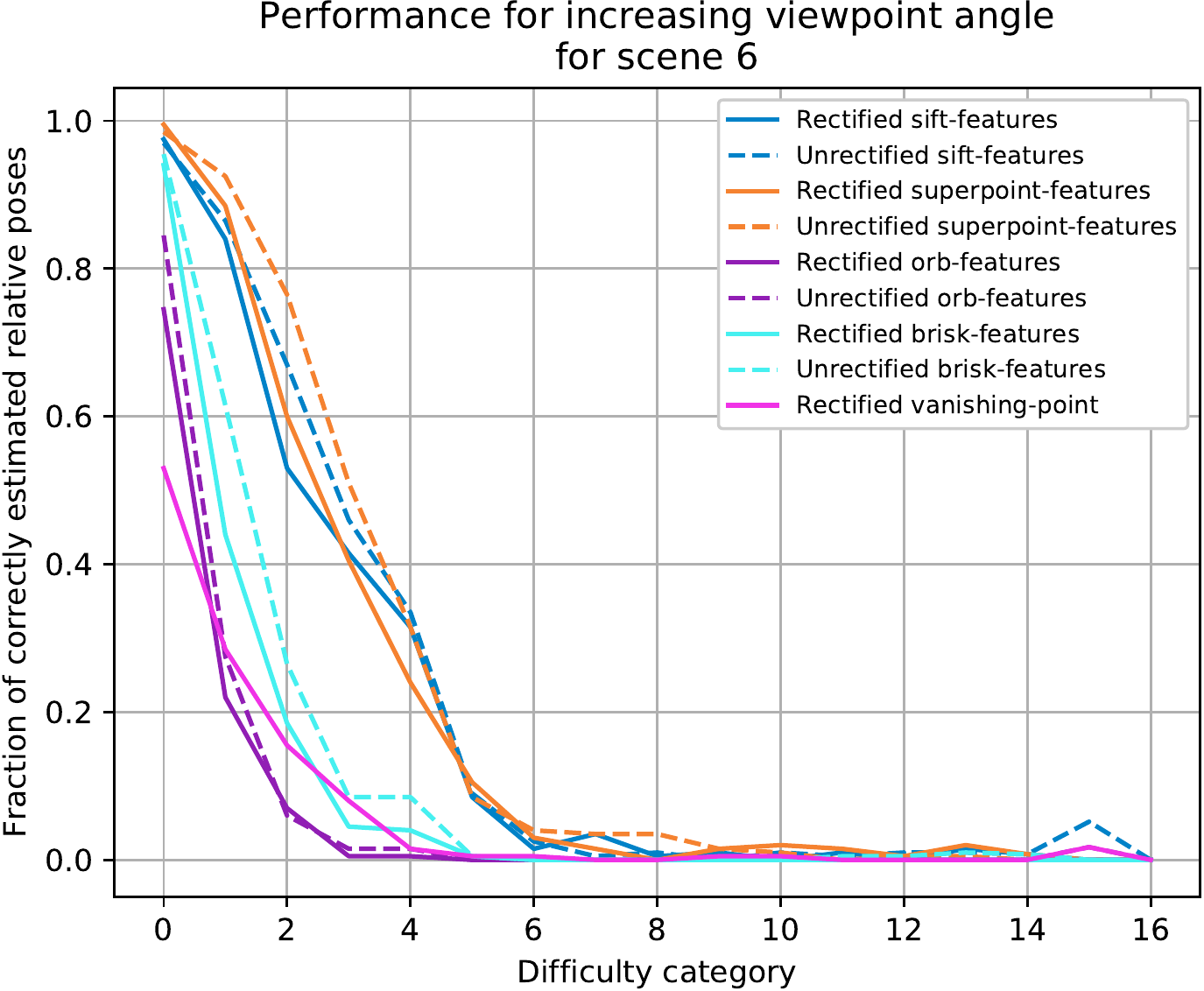}
    \caption{Performance degradation due to increasing viewpoint difference. For each of the difficulty categories (labelled from 0 to 17 on the x-axis), the y-value shows the fraction of image pairs for which the relative rotation was estimated correctly to within 5 degrees. Feature matching in the original images was compared with our rectification approach, for a variety of local features. Depth-based rectification is helpful overall, particularly for scenes with dominant planes, and more or less reduces to regular feature matching for scenes where no planes can be extracted.  \emph{Left:} Results for all images, over all scenes, in  the entire dataset.  \emph{Middle:} A scene where most image pairs show the same plane, and this plane takes up a large portion of the images. \emph{Right:} A scene containing many small facades, each often occupying a small part of the image, and some non-planar scene structures. }
    \label{fig:results:degradation}
\end{figure}

Fig. \ref{fig:results:degradation} shows the performance of feature matching directly on the image pairs, \vs matching after perspective rectification. The 18 difficulty classes as described in Sec. \ref{sec:dataset} are listed along the $x-\text{axis}$, and the fraction of image pairs successfully localized in that difficulty class is shown on the $y-\text{axis}$. 

As can be seen in the figures, extracting perspectively corrected features can improve the pose estimation performance for planar scenes, particularly for SIFT and BRISK features. Overall, SuperPoint features seem to be more robust to viewpoint changes, which is natural since the SuperPoint feature is trained by performing homographic warps of patches. The ORB features show less improvement from using perspectively corrected images. This may have to do with the fact that these are not scale-invariant, and thus only correcting for the projective distortion, but not for the scale, may not be sufficient for obtaining good feature matching performance for these features. 


In the supplementary material, localization rate graphs, like the middle and right figures in Fig. \ref{fig:results:degradation}, can be found for all eight scenes. 


\subsection{Re-localization from Opposite Viewpoints}
\label{sec:experiments:localization}
For our next experiment, we consider a re-localization scenario for autonomous driving. 
More precisely, we consider the problem of re-localizing a car driving down streets in the opposite direction from its first visit to the scene. 
Such a problem occurs, for example, during loop closure detection inside SLAM. 

\begin{figure}[t]
\begin{minipage}{0.32\columnwidth}\centering
    \includegraphics[width=\linewidth]{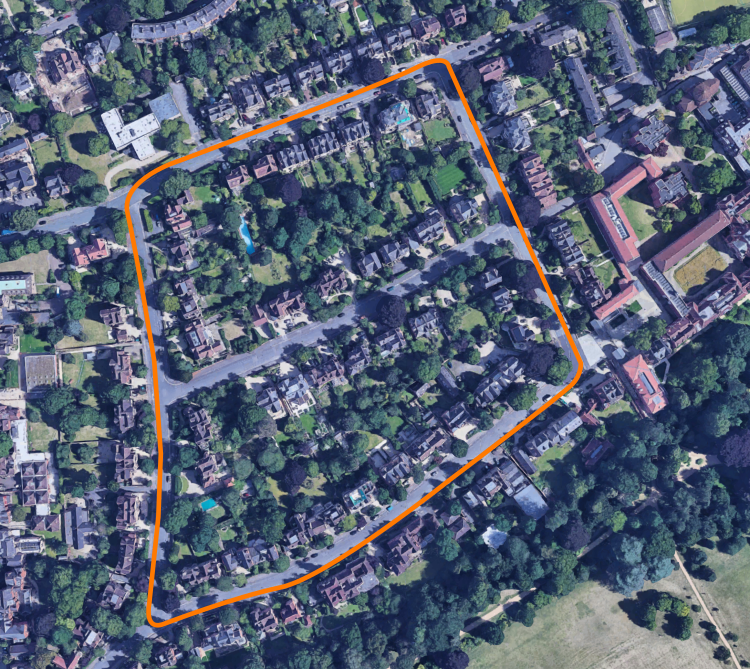}
\end{minipage}
\begin{minipage}{0.28\columnwidth}\centering
    \includegraphics[width=\linewidth]{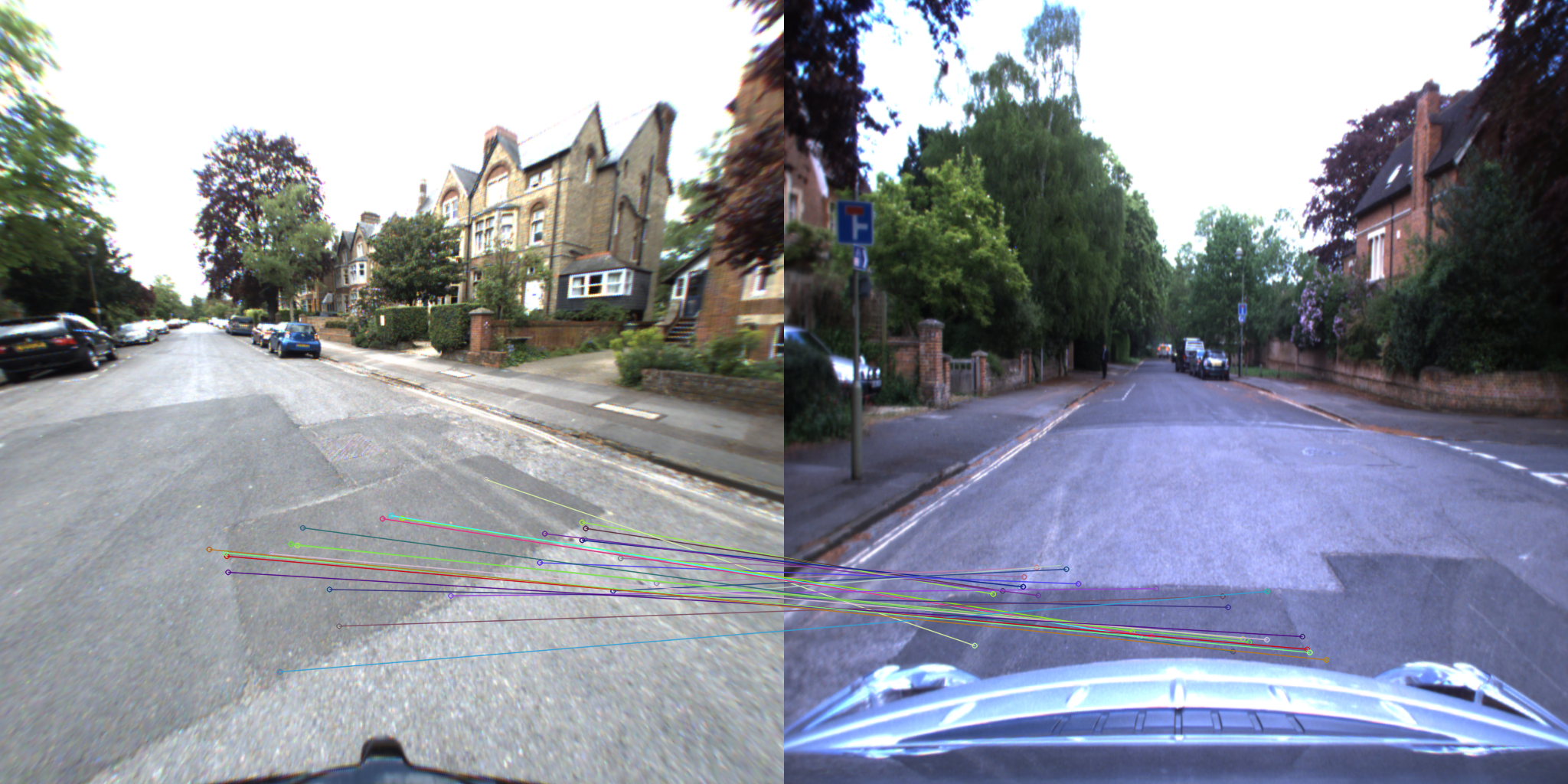}\\
    \includegraphics[width=\linewidth]{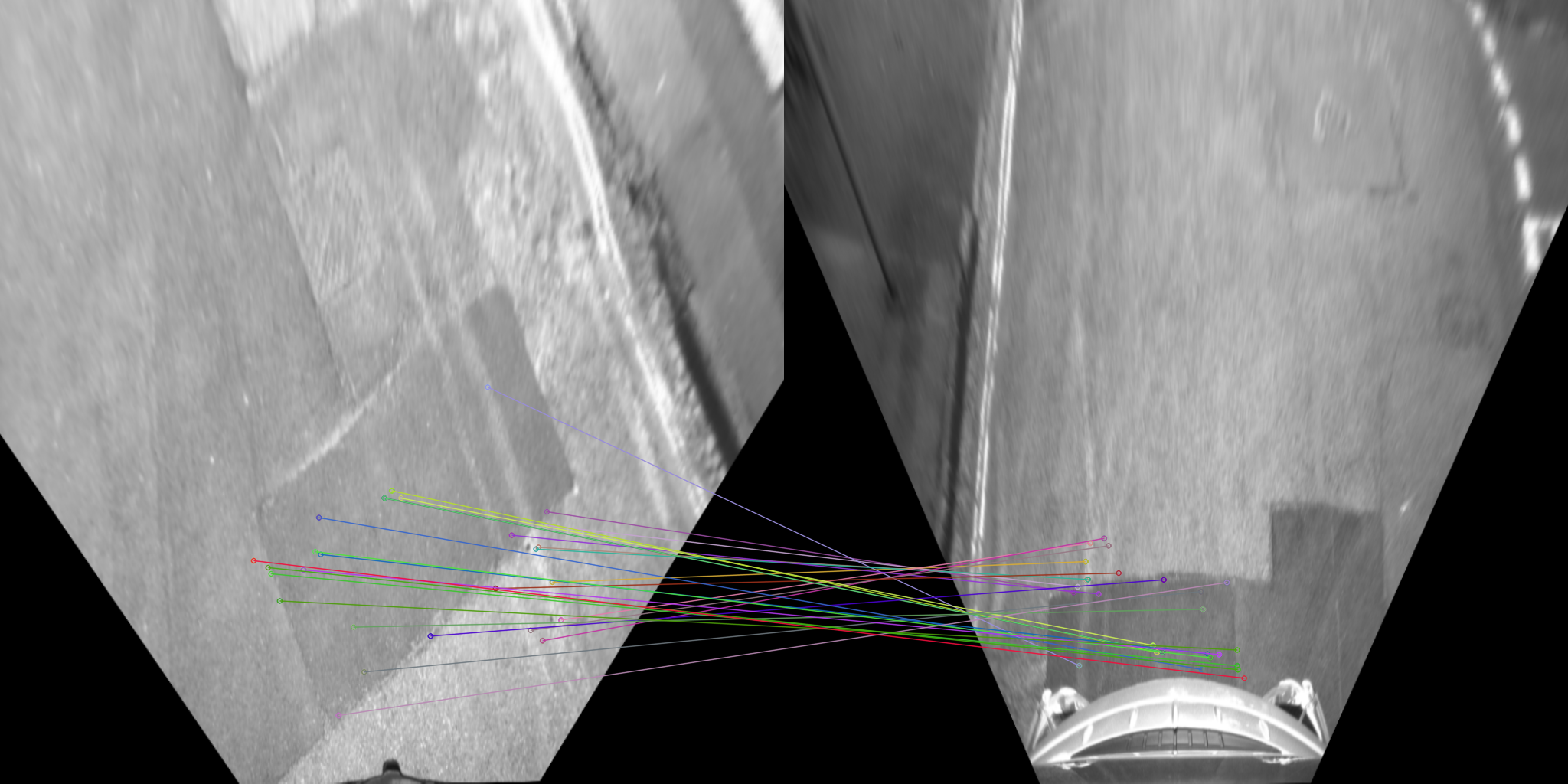}
\end{minipage}
\begin{minipage}{0.36\columnwidth}\centering
    \includegraphics[width=0.8\linewidth]{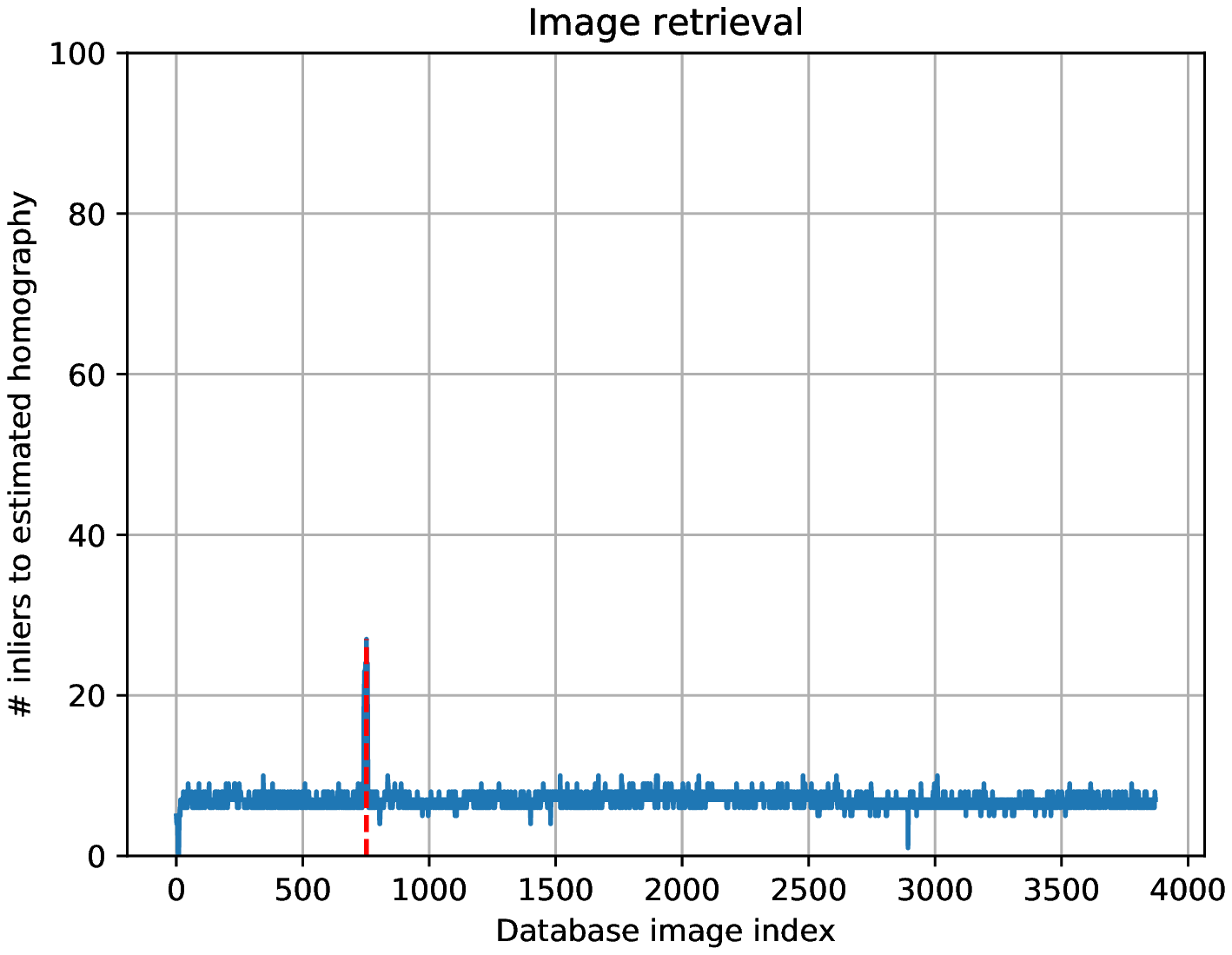}\\
    \footnotesize{(c)}\\
    \vspace{2pt}
    \begin{minipage}{\linewidth}
    \centering
    \resizebox{0.5\linewidth}{!}{
    \begin{tabular}{| c | c | c |}
        \hline
         & Seq. 1 & Seq. 2\\ \hline 
        {Ours} & \textbf{98.1} \% & \textbf{97.1} \% \\ \hline 
        Standard & 28.2 \% & - $/$ - \\ \hline 
    \end{tabular}
    }
    \end{minipage}
\end{minipage}\\
\begin{minipage}{0.32\columnwidth}\centering
\footnotesize{(a)$^*$}
\end{minipage}
\begin{minipage}{0.28\columnwidth}\centering
\footnotesize{(b)}
\end{minipage}
\begin{minipage}{0.36\columnwidth}\centering
\footnotesize{(d)}
\end{minipage}
\caption{Re-localization from Opposite Viewpoints.
(a) Satellite imagery of the area covered in the RobotCar ``Alternate Route'' dataset. We show the trajectory of one of the sequences overlaid in orange.
(b)  Example of feature matching between rear and frontal images in the RobotCar dataset. Feature matching is performed in the rectified space
(bottom), and then visualised in the original images (top). 
(c) Results of the search for the best front-facing database image for the rear-facing image from (b). 
(d) Localization results on the two sequences of the RobotCar dataset as the percentage of localized images. We compare our approach based on unwarping the ground plane with matching features in the original images.
\small{$^*$ Image taken from Google Maps. Imagery \textcopyright 2020 Google, Map data \textcopyright 2020}
}
\label{fig:results:robotcar}
\end{figure}

We use a subset of the Oxford RobotCar dataset~\cite{maddern20171}, shown in Fig.~\ref{fig:results:robotcar}, namely the ``Alternate Route'' dataset already used in~\cite{toft2017long}. 
The dataset consists of two traversals of the scene. 
We use 3,873 images captured by the front-facing camera of the RobotCar during one traversal as our database representation. 
729 images captured by the car's rear-facing camera during the same traversal as well as an additional 717 images captured by the rear camera during a second traversal (captured around 10 minutes after the first one) are used as query images. 
As a result, there is a 180$^\circ$ viewpoint change between the query and database images. 

We determine the approximate location from which the query images were taken by matching SIFT features between the query and database images. 
We compare our approach, for which we only unwarp the ground plane and use SIFT features, against a baseline that matches features between the original images\footnote{For the baseline, we only use approximately every 2nd query image.}. 
For both approaches, we use a very simple localization approach that exhaustively matches each query image against each database image. 
For our approach, we select the database image with the largest number of homography inliers, estimated using RANSAC~\cite{fischler1981random}. 
For the baseline, we select the database image with the largest number of fundamental matrix inliers, since we noticed that most correct matches are found on the buildings on the side of the road, and the corresponding points in the two images are thus not generally related by a homography. Due to the 180$^\circ$ change in viewpoint, the query and its corresponding database image might be taken multiple meters apart. 
Thus, it is impossible to use the GPS coordinates provided by the dataset for verification. 
Instead, we manually verified whether the selected database image showed the same place or not (see also the supp. video). 

Tab.~\ref{fig:results:robotcar}(d)
shows the percentage of correctly localized queries for our method and the baseline. 
As can be seen, our approach significantly ourperforms the baseline. 
A visualization for one query image and its corresponding database image found by our method is shown in Fig.~\ref{fig:results:robotcar}(b), while Fig.~\ref{fig:results:robotcar}(c) shows the number of homography inliers between this query and all database images. 
As can be seen, there is a clear peak around the correctly matching database image. 
This result is representative for most images localized by our approach (\cf the supp. video), though the number of inliers in the figure is on the lower end of what is common. 

\section{Conclusion}
The results from Sec.~\ref{sec:experiments} show that our proposed approach can significantly improve feature matching performance in real-world scenes and applications in dominantly planar scenes without a significant degradation in other environments. 
They further demonstrate that our approach is often easier to use than classical vanishing point-based approaches, which was one of the main motivations for this paper. 
Yet, our approach has its limitations. 

\PAR{Limitations.} 
Similar to vanishing point-based methods, our approach requires that the planar structures that should be undistorted occupy a large-enough part of an image. 
If these parts are largely occluded, \eg, by pedestrians, cars, or vegetation, it is unlikely that our approach is able to estimate a stable homography for unwarping. 
Further, the uncertainty of the depth predictions increases quadratically with the distance of the scene to the camera (as the volume projecting onto a single pixel grows quadratically with the distance). 
As a result, unwarping planes too far from the camera becomes unreliable. 
In contrast, it should be possible to relatively accurately undistort faraway scenes based on vanishing points or geometrically repeating elements. 
This suggests that developing hybrid approaches that adaptivley choose between different cues for perspective undistortion is an interesting avenue for future research. 

Another failure case results from the fact that all training images seen by the depth prediction network have been oriented upright. 
As such, the network fails to produce meaningful estimates for cases where the images are rotated. 
However, it will be easy to avoid such problems in many practical applications: it is often possible to observe the gravity direction through other sensors or to pre-rotate the image based on geometric cues~\cite{Xian2019ICCV}. 


\PAR{Future work.} 
We have shown that using existing neural networks for single-image depth prediction to remove perspective distortion leads to a simple yet effective approach to improve the performance of existing local features. 
A natural direction for further work is to integrate the unwarping stage into the learning process for local features. 
Rather than assuming that perspective distortion is perfectly removed, this would allow the features to compensate for inaccuracies in the undistortion process. 
Equally interesting is the question whether feature matching under strong viewpoint changes can be used as a self-supervisory signal for training single-image depth predictors: 
formulating the unwarping stage in a differentiable manner, one could use matching quality as an additional loss when training such networks.

\PAR{Acknowledgements} 
The bulk of this work was performed during an internship at Niantic, and the first author would like to thank them for hosting him during the summer of 2019. This work has also been partially supported by Swedish Foundation for Strategic Research (Semantic Mapping and Visual Navigation for Smart Robots) and the Chalmers AI Research Centre (CHAIR) (VisLocLearn). We would also like to extend our thanks Iaroslav Melekhov, who has captured some of the footage.  

\clearpage
%
%
\bibliographystyle{splncs04}
\bibliography{main_bib}
\end{document}


\pagestyle{headings}
\mainmatter
\def\ECCVSubNumber{2583}  

\title{Single-Image Depth Prediction Makes Feature Matching Easier\\
Supplementary Material}

\titlerunning{ECCV-20 submission ID \ECCVSubNumber} 
\authorrunning{ECCV-20 submission ID \ECCVSubNumber} 
\author{}
\institute{Paper ID \ECCVSubNumber}

\maketitle



In this document we present some additional results and expand on some of the topics in the main paper. Specifically, we provide results on the Aachen Day-Night dataset, which evaluates localization of nighttime query images against a 3D model build from daytime images.
 We also provide more detailed information on the MonoDepth model used and how it was trained (cf. Sec.~3.1 in the main paper), pose estimation results for eight individual scenes of the dataset (cf. Sec.~5.1 in the main paper), as well as example images from each scene (cf. Fig.~4 in the main paper), and a comparison of SIFT and SuperPoint features in the RobotCar experiments (cf. Sec.~5.2 in the main paper), as well as an evaluation on three scenes from the Extreme View Dataset. 

We also provide a supplementary video showing the performance of our approach on the RobotCar dataset.

\section{Additional Results on Aachen-Day Night}
\label{sec:aachen}

In addition to the experiments on the RobotCar dataset, we also evaluated our approach on the nighttime queries of the Aachen Day-Night dataset~\cite{Sattler2018CVPR,Sattler12BMVC}. 
We follow the experimental setup for the local feature challenge of the CVPR 2019 workshop on ``Long-Term Visual Localization under Changing Conditions": each each nighttime query image is matched against a pre-defined set of daytime database images. 
Similarly, daytime database images are matched with each other. 
The known poses and intrinsics of the database images, as well as the feature matches between them, are then used to triangulate the 3D scene structure in COLMAP~\cite{schoenberger2016sfm}. 
Finally, the matches between the nighttime queries and the database images, together with known intrinsics for the queries, are used to estimate the camera poses of the query images in COLMAP~\cite{schoenberger2016sfm}. 
We build on the code provided by the organizers\footnote{\url{https://github.com/tsattler/visuallocalizationbenchmark/tree/master/local_feature_evaluation}}, with one small difference: 
the original code performs mutual nearest neighbor matching whereas we use a Lowe ratio test~\cite{lowe2004distinctive} with a threshold of 0.8 as we observed better results when using the ratio test. 

For this experiments, we extracted SIFT features using OpenCV, both on the original images and on the rectified versions obtained by our approach. 
Following~\cite{Sattler2018CVPR}, we report the percentage of query images localized within (0.5m, 2$^\circ$), (1m, 5$^\circ$), and (5m, 10$^\circ$) of the reference pose (using the evaluation server provided at \url{https://www.visuallocalization.net/}. 
Using the original images, we obtain 23.5\%, 35.7\%, and 48.0\%, respectively. 
Extracting features on images rectified by our approach improves the performance to 26.5\%, 40.8\%, and 53.1\%, respectively. 
As can be seen, our approach is able to significantly improve localization performance. 
This clearly shows that removing perspective distortion before feature extraction improves pose estimation accuracy under changing viewpoints. Furthermore the results indicate that our method does not cause degradation despite challenges such as day-night changes.

\section{Our Depth Prediction Network}
\label{sec:our-network}
In this section we expand on the singe-image depth prediction network used in our method.

\begin{figure}
    \centering
    \includegraphics[width=\linewidth]{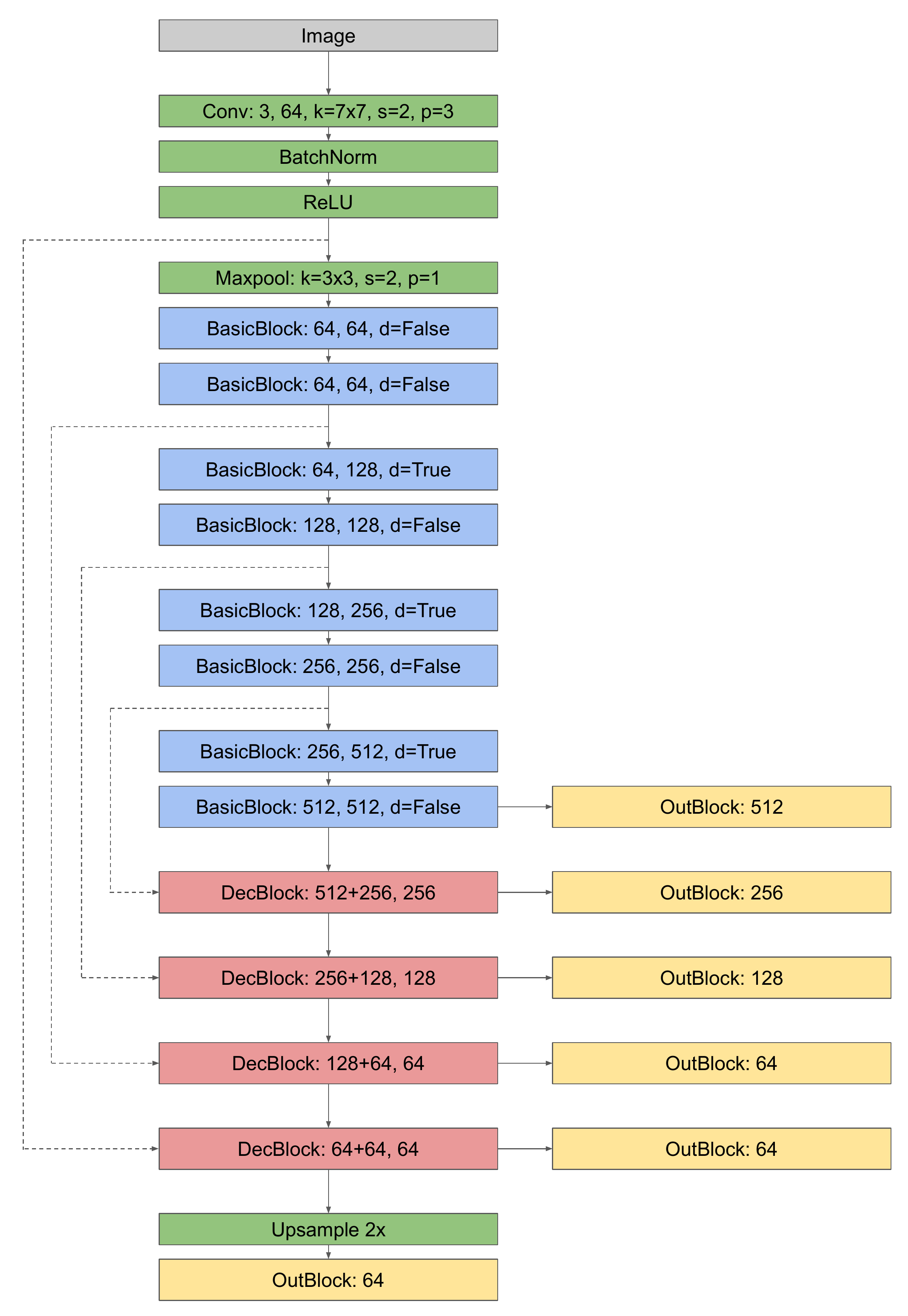}
    \caption{Architecture of our network. Please see Figure~\ref{fig:network:buildingblocks} for details on building blocks.}
    \label{fig:network:architecture}
\end{figure}

\PAR{Architecture}
The network architecture is a U-Net similar to the Resnet18-based architecture in Monodepth2~\cite{monodepth2}, but with double convolutions in the decoder.
Please see Figure~\ref{fig:network:architecture} for a visualization of the network architecture used.

\begin{figure}
    \centering
    \includegraphics[width=\linewidth]{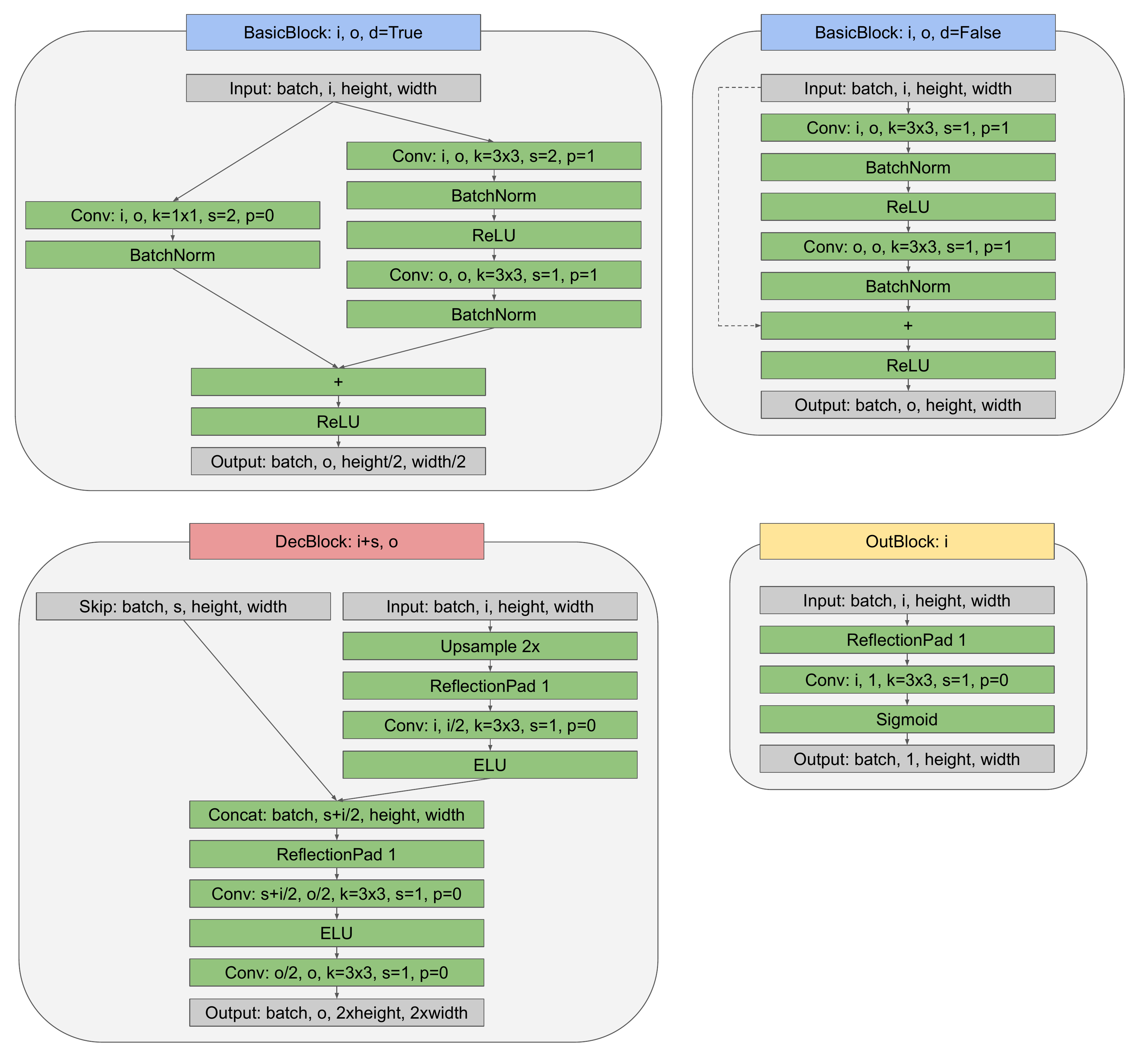}
    \caption{Building blocks used in the architecture (Figure~\ref{fig:network:architecture}) of our network.}
    \label{fig:network:buildingblocks}
\end{figure}

\PAR{Training}
We trained our network with several datasets: Our own stereo video footage, Megadepth~\cite{megadepth}, and Matterport~\cite{matterport}. The network was trained with a $512 \times 256$ resolution as input (similarly $256 \times 512$ for portrait data).

We scale the sigmoid prediction of the network to be in the range $(0.5, 100)$ meters.

\PAR{Stereo data}
Our stereo data consists of several hours of stereo video captured in one European city and three US cities. The footage was captured with a landscape orientation of the cameras as well as a portrait orientation of the cameras. The cameras were calibrated so that the network predictions are metric. The cameras were re-calibrated at each capture session. 

The network was trained with the Depth Hints loss~\cite{depth-hints} on stereo data in addition to a Monodepth2 reprojection-based loss and a sky segmentation prior (see below). 
However, our results in the paper for Robotcar dataset (only) used a network that was trained without the Depth Hints loss and instead used a Monodepth2 reprojection-based SSIM+L1 loss for the training loss for the stereo data.

\PAR{Megadepth}
Megadepth~\cite{megadepth} has depth estimates that are scale-ambiguous. So, we use a scale-invariant loss (Equation 2 in~\cite{megadepth}) for the images with dense depth estimates in Megadepth.
The images that have ordinal labels are also used with a robust ordinal depth loss (equation 4 in~\cite{megadepth}).
During training the images and depth maps were cropped to the target aspect ratio $512/256$ or $256/512$ (randomly chosen as landscape or portrait) and isotropically scaled to $512 \times 256$ to be fed as input to the network.

\PAR{Matterport}
The Matterport dataset provides images with metric depth captured with Kinect-like cameras. We follow~\cite{Hu2018Revisiting} for supervised training from Matterport data, using the loss function $\log(1 + |d - t|)$, where $d$ is the network prediction and $t$ is the target depth (Equation 3 in~\cite{Hu2018Revisiting}) as well as a depth gradient loss (Equation 4).
Similarly to the Megadepth dataset, we crop and scale images during training.

\PAR{Sky loss}
We also trained a segmentation network using the ADE20K dataset~\cite{zhou2017scene} that predicts if the pixels belongs to the sky or not. During training we use the predicted sky segmentation mask to have a small regualarization loss (weight 0.04) that forces masked pixels to have maximum depth (100 meters in our model) with L1 loss on depth values.


\section{Robotcar with Superpoint}
In this section we elaborate and motivate more on the choice of SIFT features for the RobotCar experiments. One of the main reasons for using SIFT is its invariance to in-plane rotations, a property not possessed by the SuperPoint or D2-Net features. 
This rotational invariance is crucial to the presented localization experiments, since unlike in an upright photo, there is no clear preferred direction in a top-down view of the road. We may thus expect the rectified query and database images to have any possible relative rotation. 

SuperPoint features are trained by applying homographic warps to patches to obtain correspondences. These warps include rotations, but the publicly available model has been trained on only small rotations, leading to a reduced robustness to rotations. 
In this section we present an experiment that demonstrates this, illustrating that there are still some applications where SIFT continues to be an appropriate choice. 

For pairwise matching, the same procedure is followed as in the main paper: features are extracted from the rectified patches, and features close to the warped image border are discarded. Pairwise matching is performed between the images using approximate nearest neighbour matching~\cite{muja2009flann}, and the obtained matches are then geometrically verified by fitting a homography to them using RANSAC~\cite{fischler1981random} with a 10 pixel inlier threshold. Lastly, the number of inliers to the homography is saved for this query. 
Specifically, we go through each of the 729 query images in the first sequence of the RobotCar dataset used in the main paper. For each query image, we retrieve the top-ranked database image from the experiments in Sec. 5.2 of the main paper, and we check whether SuperPoint is able to establish matches between these images. Since the image retrieval failed for a few images, we do not expect all of these query-database image pairs to match. However, since the success rate was larger than 98\% for this dataset, performing pairwise feature matching between the query image and the top retrieved database image should indicate whether or not SuperPoint features are suitable for this task at all. 
\begin{figure}[t]
    \centering
    \includegraphics[width=0.9\linewidth]{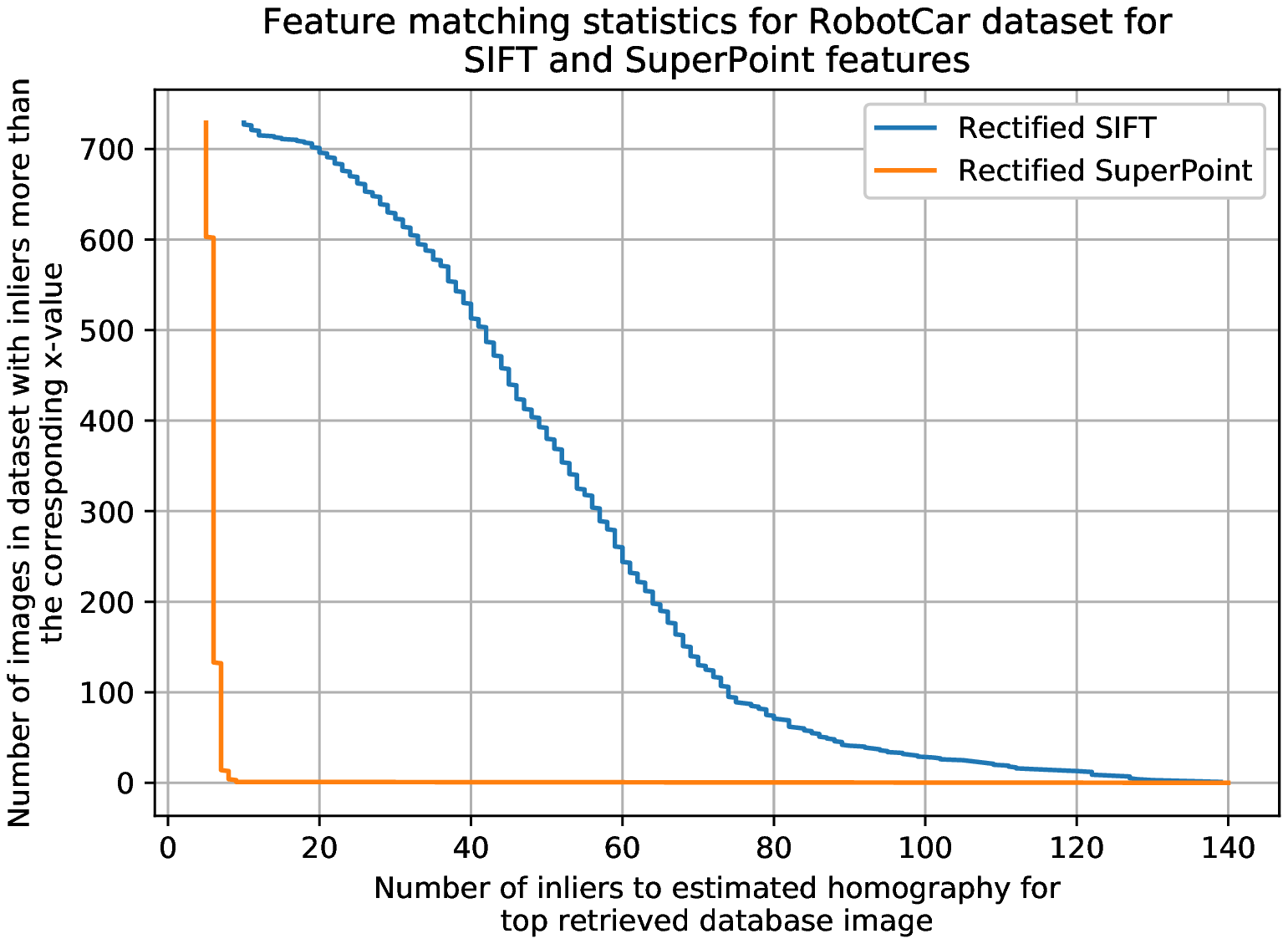}
    \caption{Number of inliers to the homography estimated during pairwise matching between the query images and the top-retrieved database images for the first sequence of the RobotCar dataset. The $y$-values denote the number of images whose homography has at least the number of inliers specified on the $x$-axis. }
    \label{fig:robotcar:sp-inliers}
\end{figure}

Fig. \ref{fig:robotcar:sp-inliers} shows the number of inliers to the estimated homography from both the SIFT matching, as well as the SuperPoint matching. For each value on the $x$-axis, the corresponding $y$-value shows the number of query images (out of the 729) whose final estimated homography had that number of inliers or more. A ''higher'' curve is thus better. 

As expected, due to the rotational variance of SuperPoint, it fails to reliably match essentially all image pairs. No query image had more than nine inliers to the estimated homography. 

\section{Results with and without enforcing orthogonal normals during clustering} 
\begin{figure}[t]
    \centering
    \includegraphics[width=0.9\linewidth]{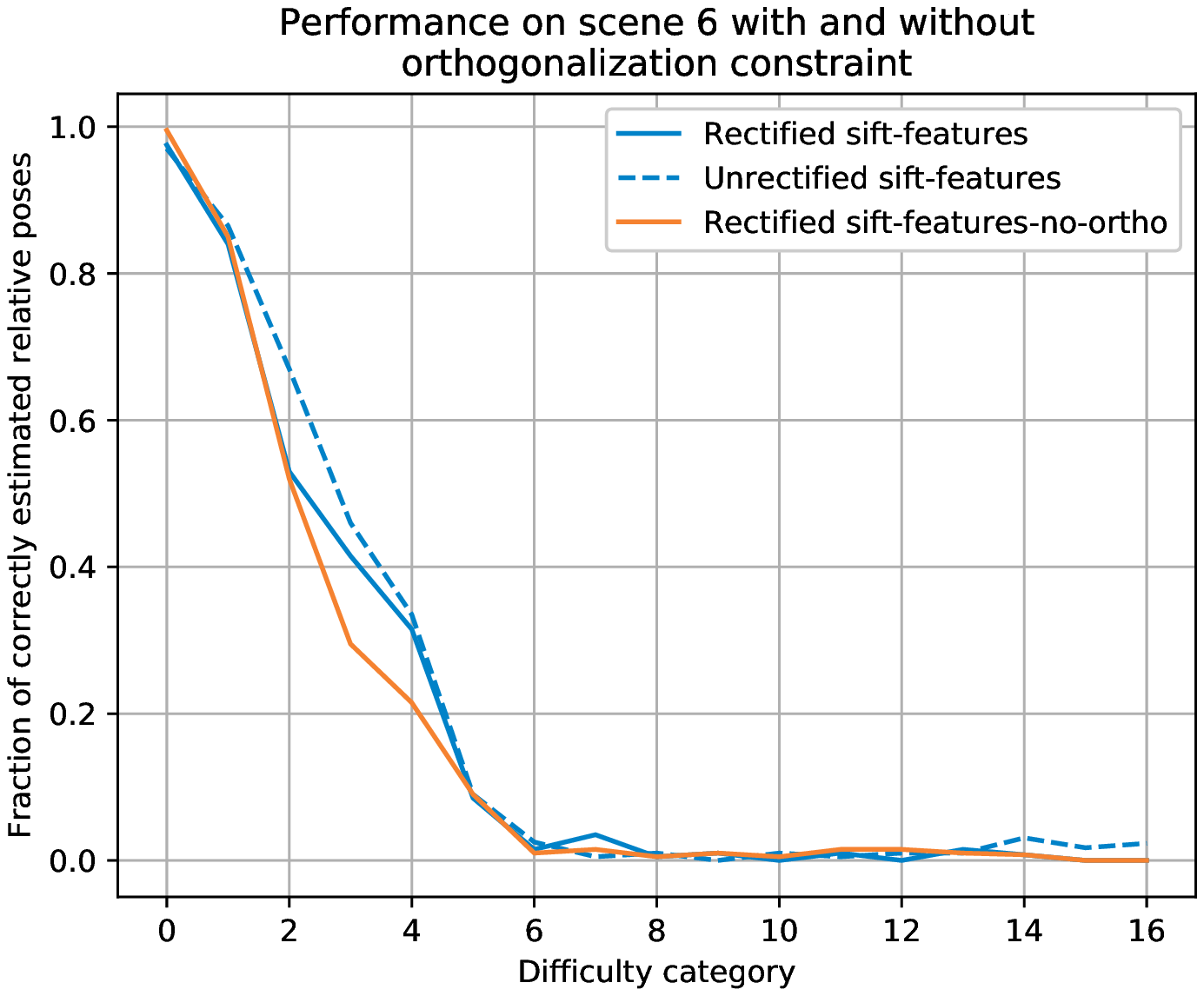}
    \caption{Performance on scene 6 with and without enforcing the normal clusters to be orthogonal. } 
    \label{fig:no-ortho-normals}
\end{figure}

Experiments were performed on scene 6 of our dataset when not enforcing orthogonality between the normal clusters. Instead, planes were found by histogramming the normals into 200 bins on the unit sphere. Thresholding and non-maximum suppression were then performed to obtain a set of plane hypotheses. Otherwise the pipeline was the same as in the main experiments. Results using this method is shown in Fig. \ref{fig:no-ortho-normals}. 

As can be seen in the figure, enforcing the orthogonality improves the performance. The decreased performance without rectification is most likely due to inaccuracies in the monocular depth estimation network. Enforcing orthogonality is thus a way to reduce the noise in the depth predictions. 

\section{Performance using different monocular depth estimation networks}
\begin{figure}
    \centering
    \includegraphics[scale=0.75]{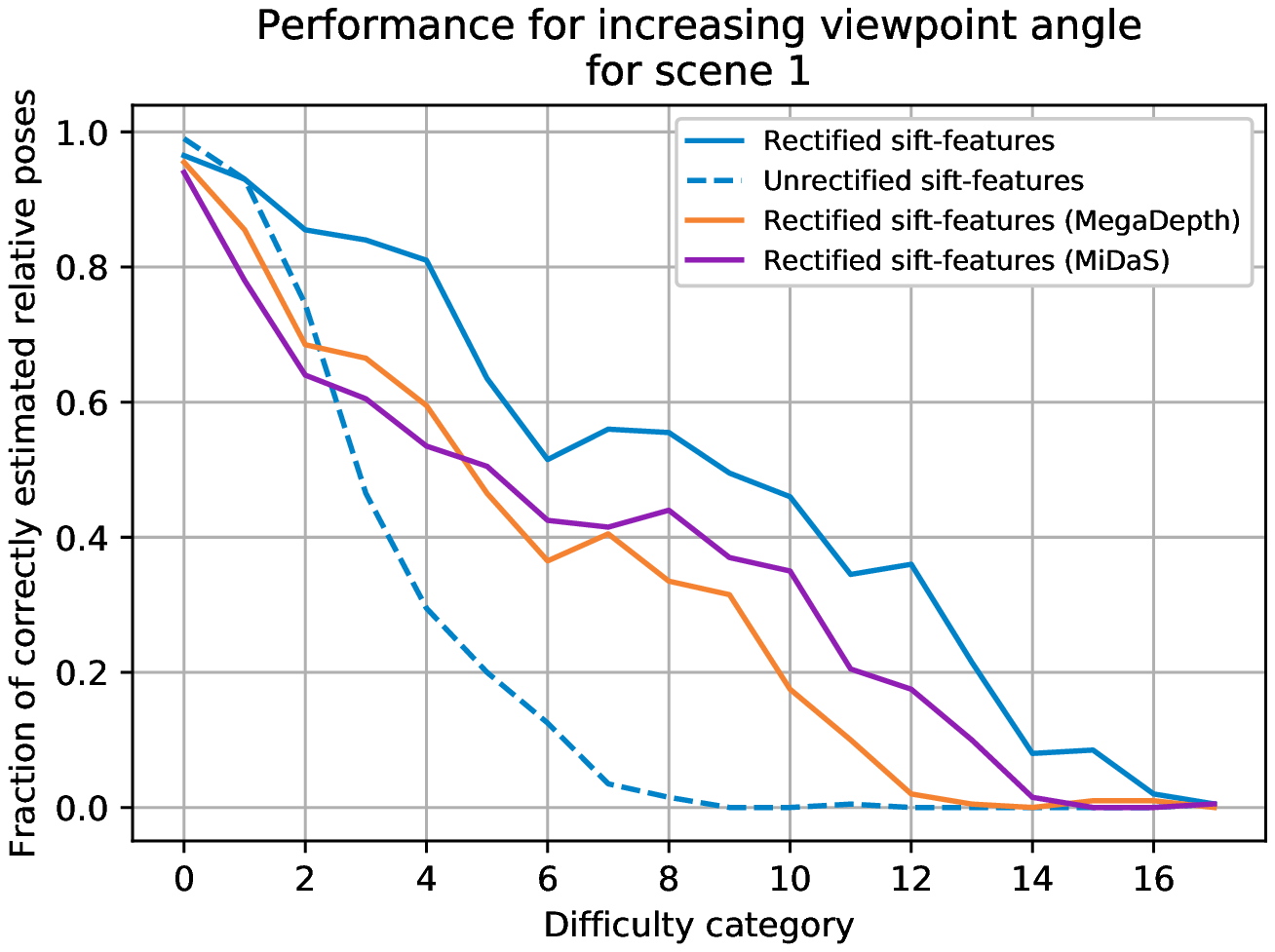}
    \caption{Performance on scene 1 using two other monocular depth prediction networks. Only the depth prediction network has changed, the rest of the pipeline remains unchanged. }
    \label{fig:different_networks}
\end{figure}

Fig. \ref{fig:different_networks} shows the results when replacing the depth prediction network used in the main paper (described in Sec. \ref{sec:our-network}) with the MegaDepth network \cite{megadepth} and MiDaS \cite{lasinger2019towards}. 

For both MiDaS and MegaDepth, we used the official implementations available on the project webpages. In MiDaS, the images are rescaled such that their largest axis equals 384, and the smaller axis is chosen as the multiple of 32 that best preserves the aspect ratio of the original image. For MegaDepth, we similarly rescale the images to have a maximum dimension of 512, with the other dimension chosen as the multiple of 32 that best preserves the original aspect ratio. 

The reason MonoDepth performs better on this scene seems to be that MiDaS and MegaDepth sometimes have difficulty separating a building facade and a cloudy gray sky, whereas the MonoDepth network does not seem to have trouble distinguishing between these. This leads to noisier estimates of the surface normal of the plane. This may perhaps be attributed to the different training data the three networks have been trained on. 

\section{Detailed results on all scenes of our dataset}
\begin{figure}
\centering
\includegraphics[width=0.39\linewidth]{figures/suppmat/individual_plots/scene1_difficulty_results.pdf}
\includegraphics[width=0.39\linewidth]{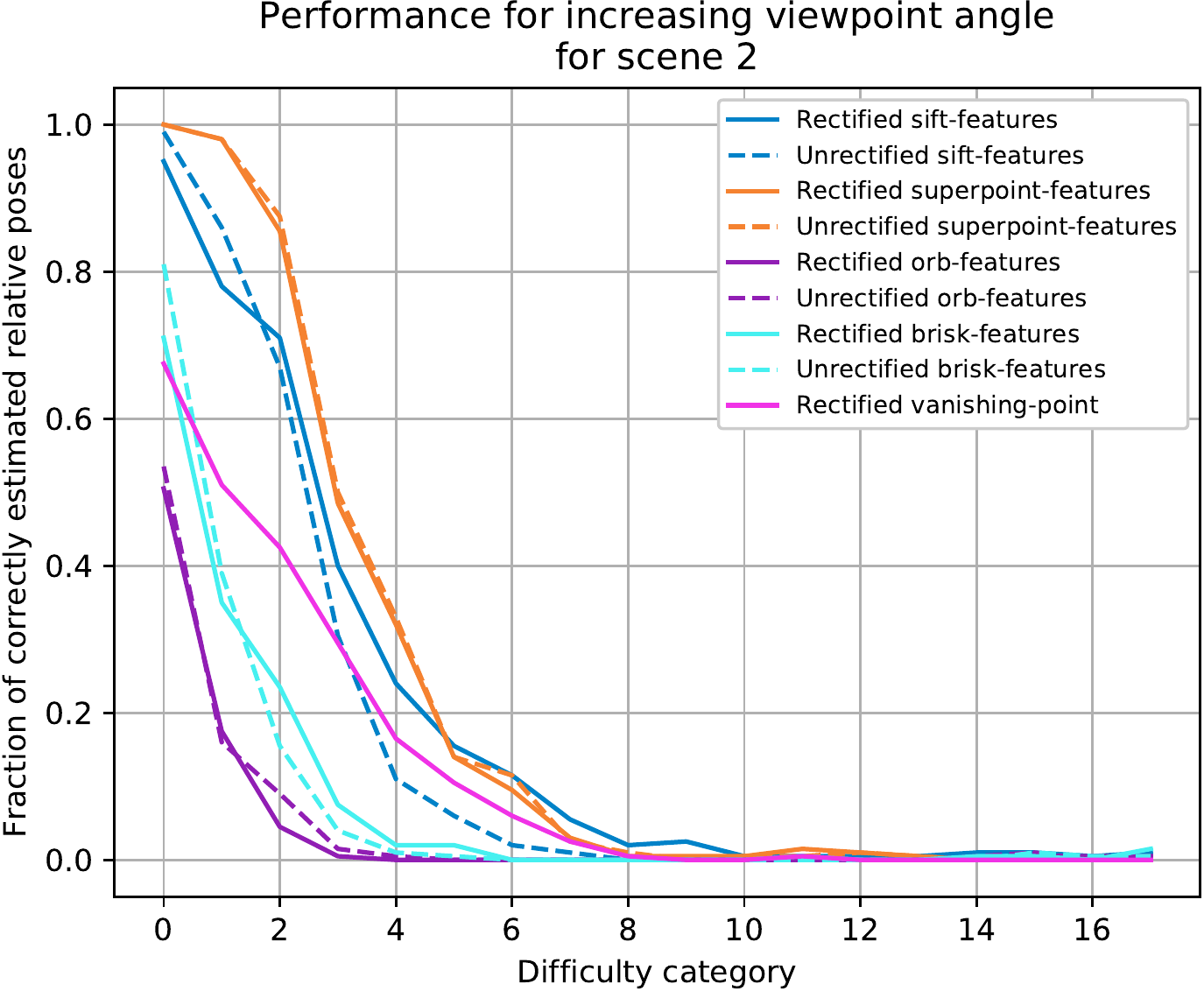}
\includegraphics[width=0.39\linewidth]{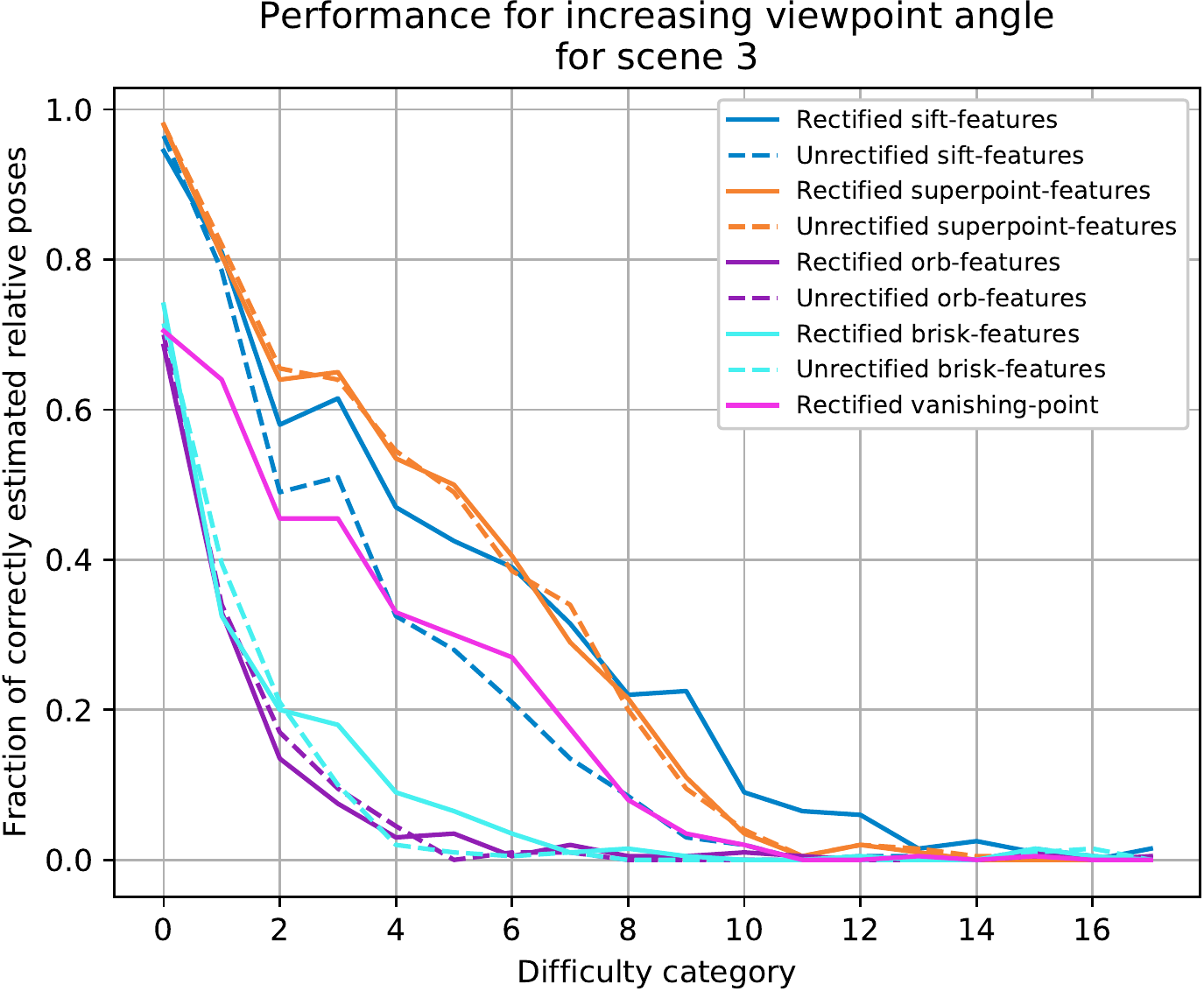}
\includegraphics[width=0.39\linewidth]{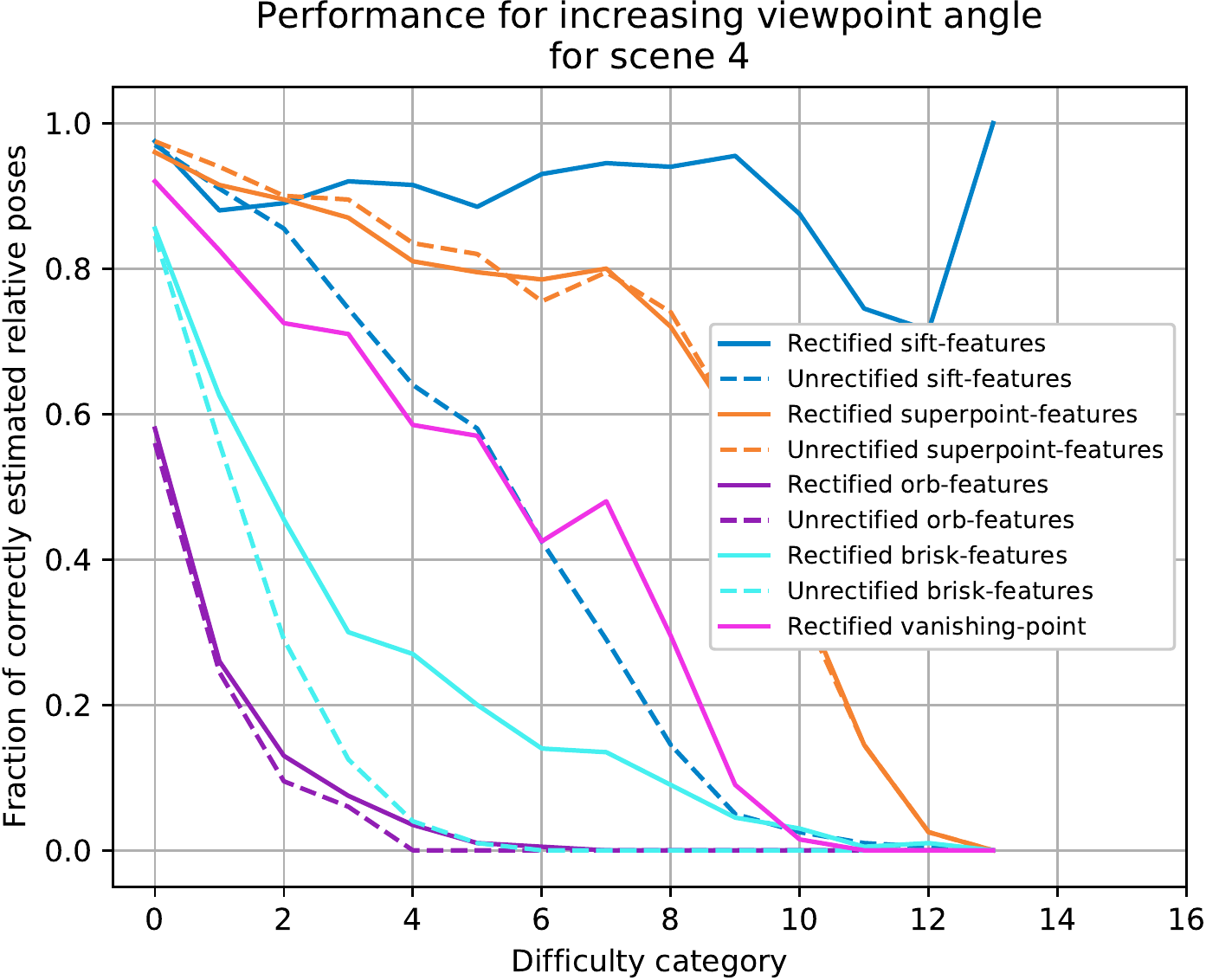} \\
\includegraphics[width=0.39\linewidth]{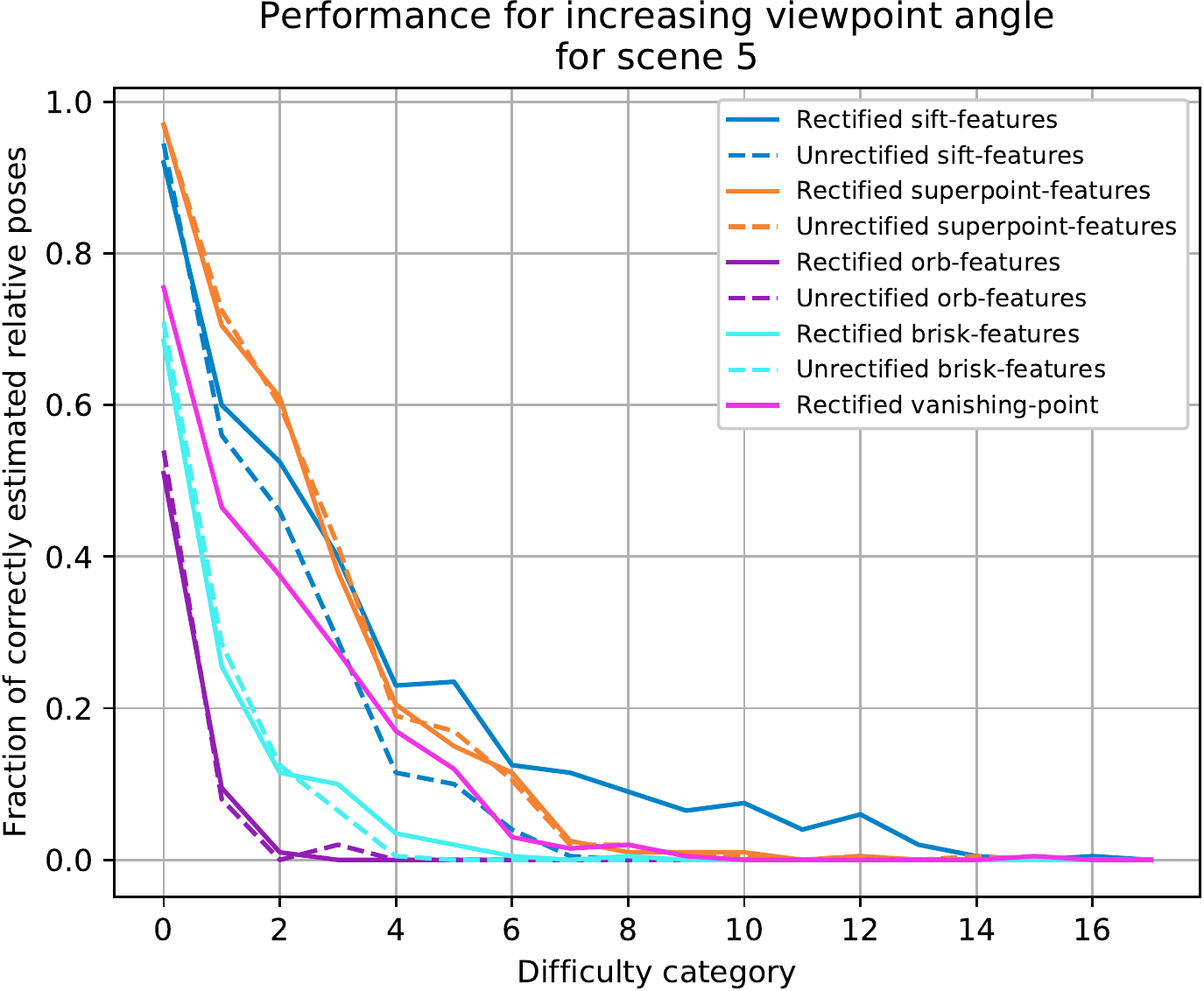}
\includegraphics[width=0.39\linewidth]{figures/suppmat/individual_plots/scene6_difficulty_results.pdf}
\includegraphics[width=0.39\linewidth]{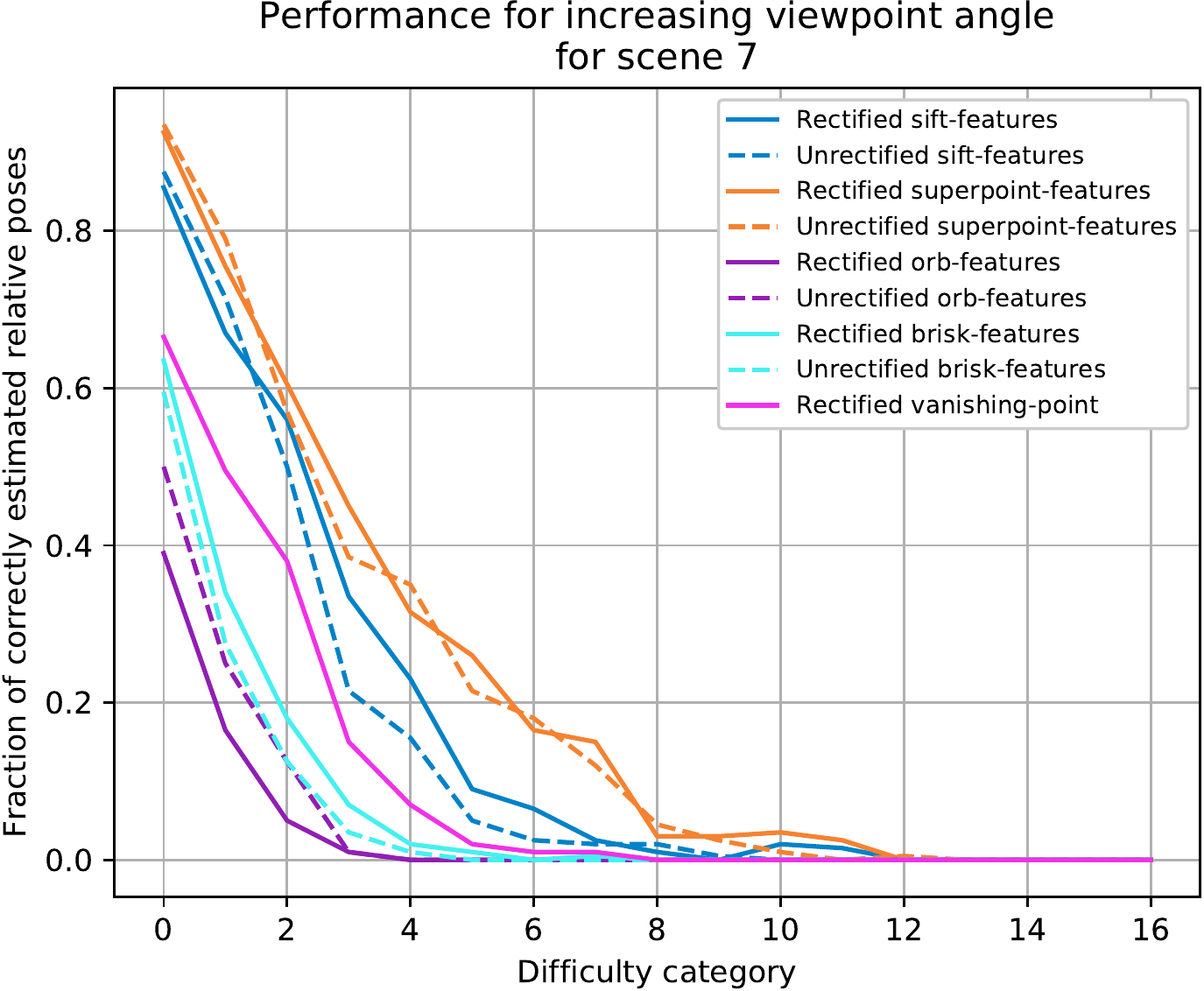}
\includegraphics[width=0.39\linewidth]{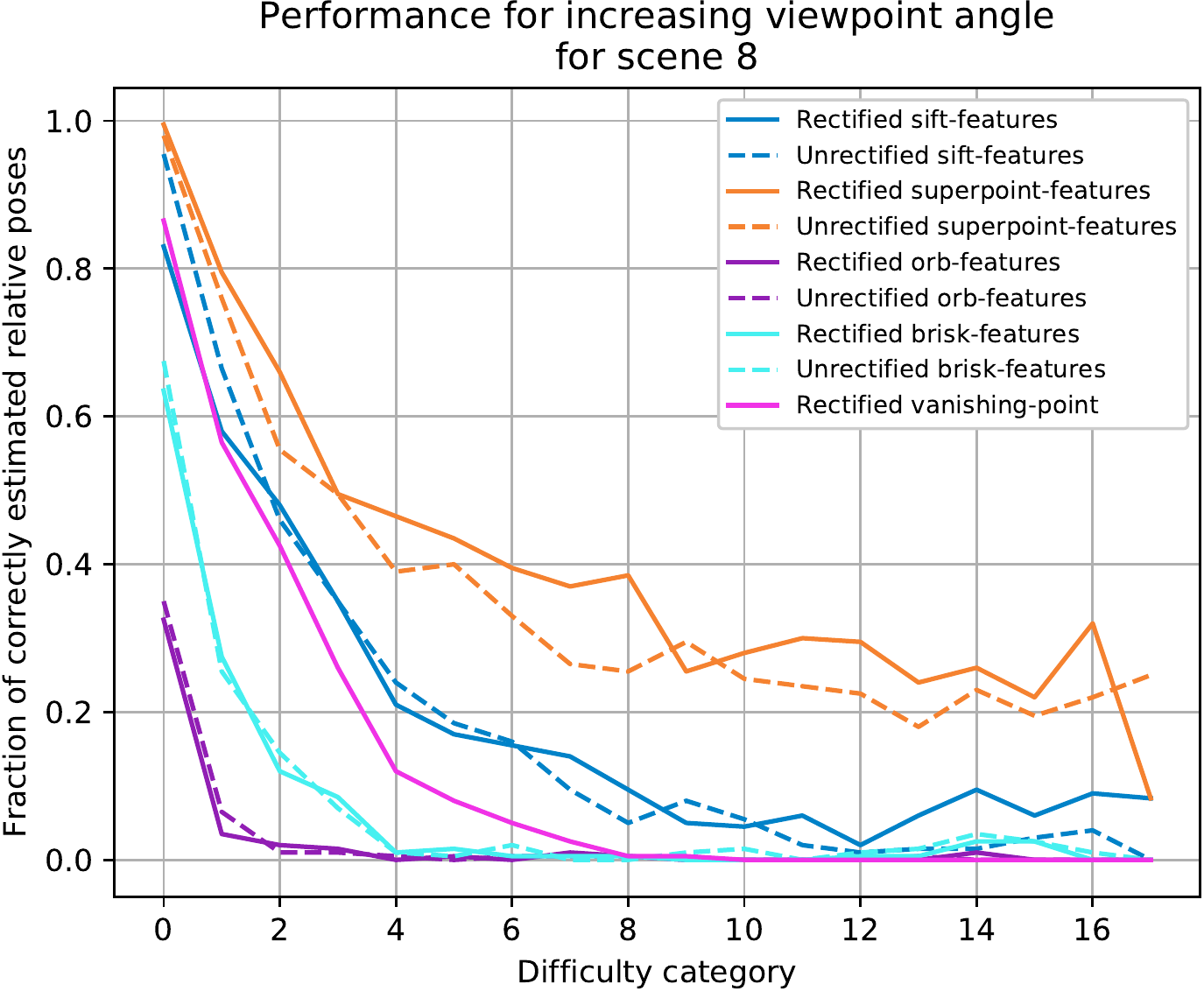}
\caption{Detailed results on the presented local feature matching dataset, showing the performance on each scene individually. }
\label{fig:all-results} 
\end{figure}

\begin{figure}
\centering
\includegraphics[width=0.115\linewidth]{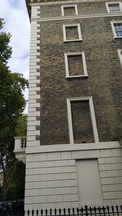}
\includegraphics[width=0.115\linewidth]{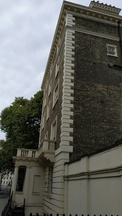}
\includegraphics[width=0.115\linewidth]{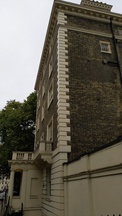}
\includegraphics[width=0.115\linewidth]{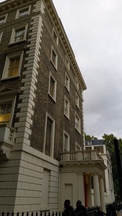}
\includegraphics[width=0.115\linewidth]{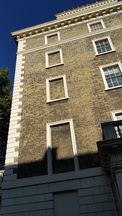}
\includegraphics[width=0.115\linewidth]{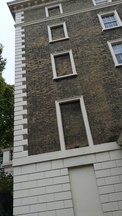}
\includegraphics[width=0.115\linewidth]{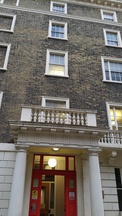}
\includegraphics[width=0.115\linewidth]{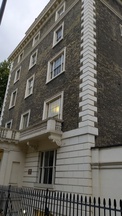} \\
\includegraphics[width=0.115\linewidth]{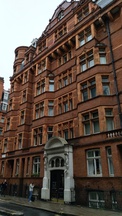}
\includegraphics[width=0.115\linewidth]{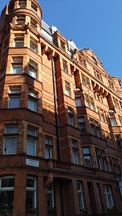}
\includegraphics[width=0.115\linewidth]{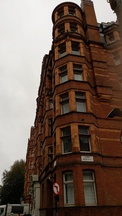}
\includegraphics[width=0.115\linewidth]{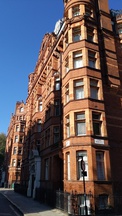}
\includegraphics[width=0.115\linewidth]{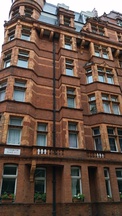}
\includegraphics[width=0.115\linewidth]{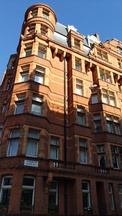}
\includegraphics[width=0.115\linewidth]{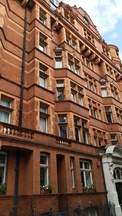}
\includegraphics[width=0.115\linewidth]{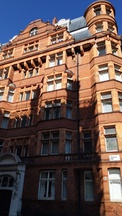} \\
\includegraphics[width=0.115\linewidth]{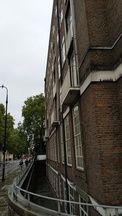}
\includegraphics[width=0.115\linewidth]{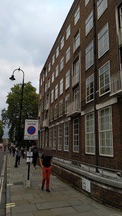}
\includegraphics[width=0.115\linewidth]{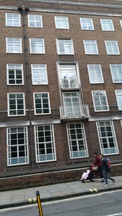}
\includegraphics[width=0.115\linewidth]{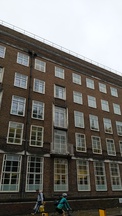}
\includegraphics[width=0.115\linewidth]{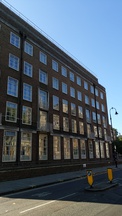}
\includegraphics[width=0.115\linewidth]{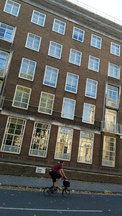}
\includegraphics[width=0.115\linewidth]{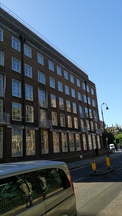}
\includegraphics[width=0.115\linewidth]{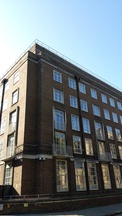} \\ 
\includegraphics[width=0.115\linewidth]{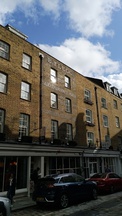}
\includegraphics[width=0.115\linewidth]{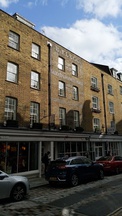}
\includegraphics[width=0.115\linewidth]{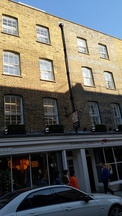}
\includegraphics[width=0.115\linewidth]{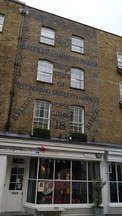}
\includegraphics[width=0.115\linewidth]{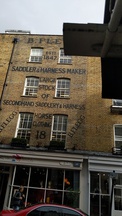}
\includegraphics[width=0.115\linewidth]{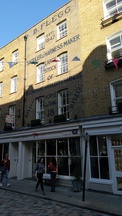}
\includegraphics[width=0.115\linewidth]{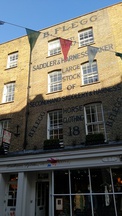}
\includegraphics[width=0.115\linewidth]{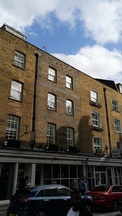} \\
\includegraphics[width=0.115\linewidth]{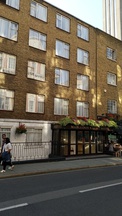}
\includegraphics[width=0.115\linewidth]{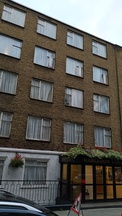}
\includegraphics[width=0.115\linewidth]{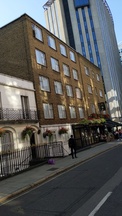}
\includegraphics[width=0.115\linewidth]{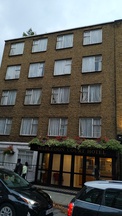}
\includegraphics[width=0.115\linewidth]{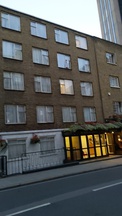}
\includegraphics[width=0.115\linewidth]{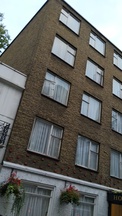}
\includegraphics[width=0.115\linewidth]{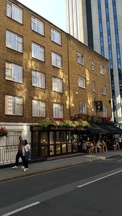}
\includegraphics[width=0.115\linewidth]{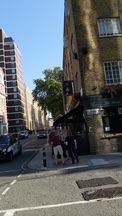} \\
\caption{Example images from our dataset for Strong Viewpoint Changes. Each row shows a sample of images showing scenes 1 to 5. }
\label{fig:example_images}
\end{figure}

\begin{figure}
\centering 
\includegraphics[width=0.24\linewidth]{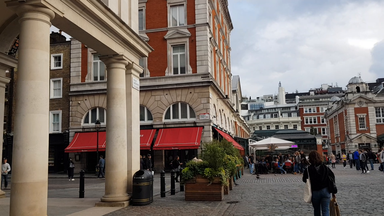}
\includegraphics[width=0.24\linewidth]{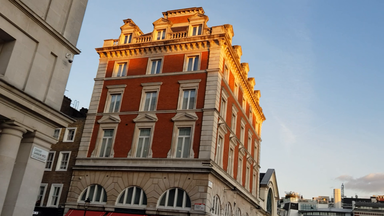}
\includegraphics[width=0.24\linewidth]{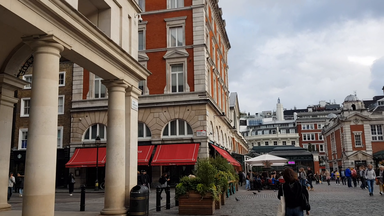}
\includegraphics[width=0.24\linewidth]{figures/suppmat/example_images/resized_scene6_ex_3.png} \\
\includegraphics[width=0.24\linewidth]{figures/suppmat/example_images/resized_scene6_ex_4.png}
\includegraphics[width=0.24\linewidth]{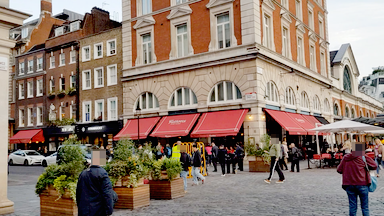}
\includegraphics[width=0.24\linewidth]{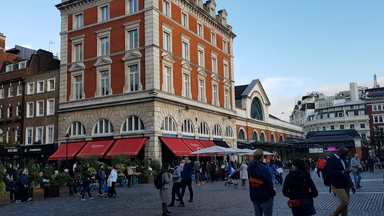}
\includegraphics[width=0.24\linewidth]{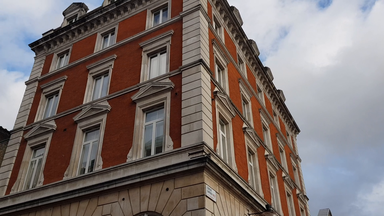} \\
\includegraphics[width=0.115\linewidth]{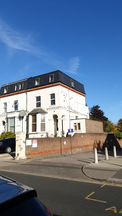}
\includegraphics[width=0.115\linewidth]{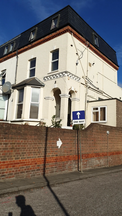}
\includegraphics[width=0.115\linewidth]{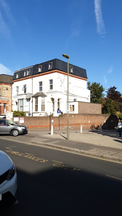}
\includegraphics[width=0.115\linewidth]{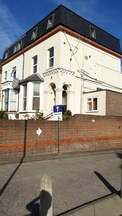}
\includegraphics[width=0.115\linewidth]{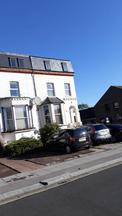}
\includegraphics[width=0.115\linewidth]{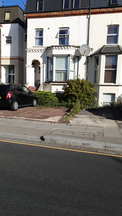}
\includegraphics[width=0.115\linewidth]{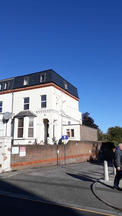}
\includegraphics[width=0.115\linewidth]{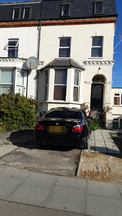} \\ 
\includegraphics[width=0.115\linewidth]{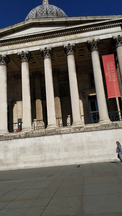}
\includegraphics[width=0.115\linewidth]{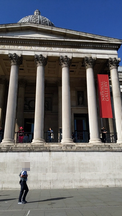}
\includegraphics[width=0.115\linewidth]{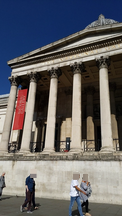}
\includegraphics[width=0.115\linewidth]{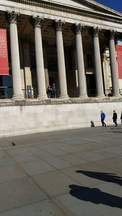}
\includegraphics[width=0.115\linewidth]{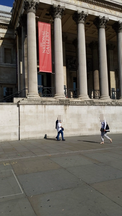}
\includegraphics[width=0.115\linewidth]{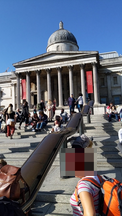}
\includegraphics[width=0.115\linewidth]{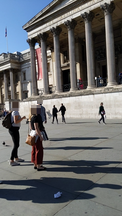}
\includegraphics[width=0.115\linewidth]{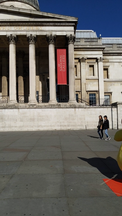} 
\caption{Example images from our dataset for Strong Viewpoint Changes. Top two rows show a sample of images for scene 6.
The following rows show images of scene 7 and 8, respectively.
}
\label{fig:example_images_2}
\end{figure}

Fig. \ref{fig:all-results} presents individual results for each of the eight scenes in the dataset. 
Figs. \ref{fig:example_images} and \ref{fig:example_images_2} show example images from each of the eight scenes in our dataset. 
Note that our dataset contains scenes of varying difficulty for our approach, ranging from scenes dominated by a planar surface (scenes 1, 3, 4, 5), roughly planar scenes (scene 2), over scenes with multiple planar surfaces (scenes 6, 7), to scenes with little dominant planes (scene 8).

We note that the proposed method of extracting features from rectified patches achieves the best performance for the datasets where a large portion of the image is taken up by one dominant plane, and the performance seems to drop as the viewed planes become smaller. This is likely due to the estimated normals getting more noisy, leading to less accurate rectifications. Since all normals assigned to a given plane are used to estimate the plane normal, fewer pixels per plane lead to fewer measurements of the plane normal, and thus a more noisy estimate. 

As a result, our method performs the best on scenes 1 to 5, where an estimate of the plane normal can be extracted fairly reliably, whereas for example in scene 8, where there are very few planar surfaces to rectify, the method more or less reduces to  feature matching using regular features. 

The performance on scene 4 is especially good. This is most likely due to the depth predictions being very accurate: some of the data used to train the depth network was captured from the surrounding areas (though none of the images in the scene have been seen during training), which may result in more accurate depths for this scene. The results may thus be indicative of what might be achieved as monocular depth estimation networks get better. 

\section{Example normal clusterings}
\begin{figure}[t]
    \centering
    \includegraphics[width = 0.24\linewidth]{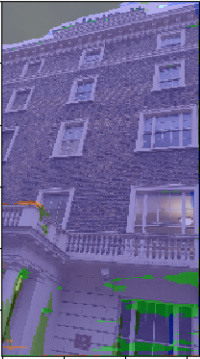} 
    \includegraphics[width = 0.24\linewidth]{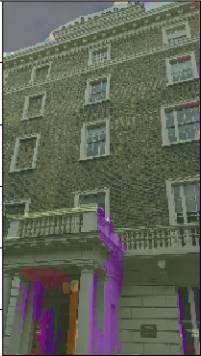} 
    \includegraphics[width = 0.24\linewidth]{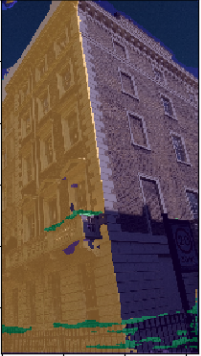} 
    \includegraphics[width = 0.24\linewidth]{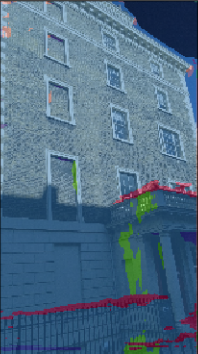} \\
    \includegraphics[width = 0.24\linewidth]{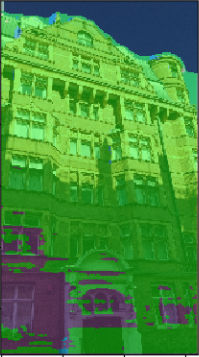} 
    \includegraphics[width = 0.24\linewidth]{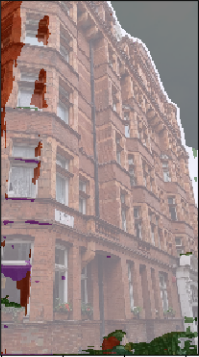} 
    \includegraphics[width = 0.24\linewidth]{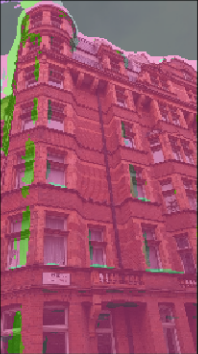} 
    \includegraphics[width = 0.24\linewidth]{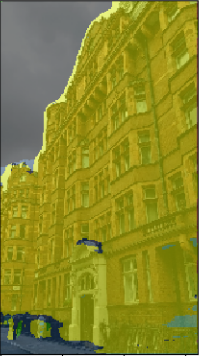} \\
    \includegraphics[width = 0.48\linewidth]{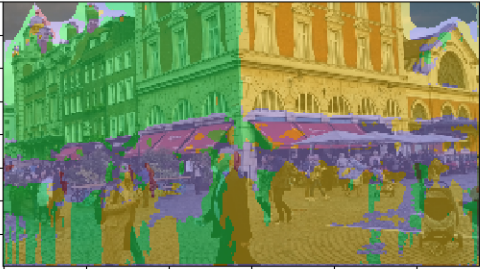} 
    \includegraphics[width = 0.48\linewidth]{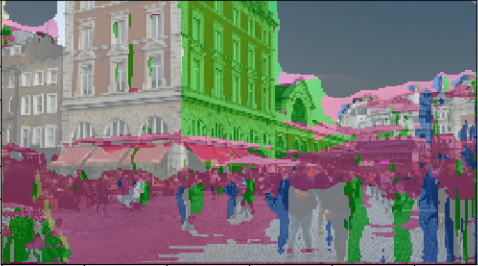} \\
    \includegraphics[width = 0.48\linewidth]{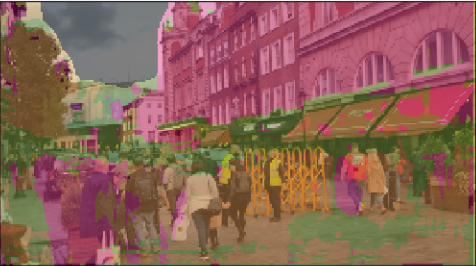} 
    \includegraphics[width = 0.48\linewidth]{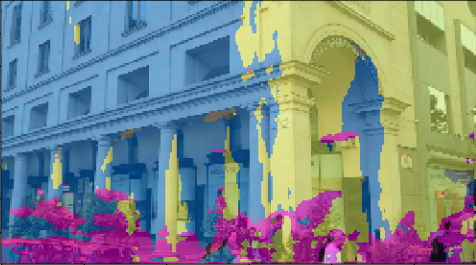} 
    \caption{Normal clustering results on four images from three of the scenes in our dataset. }
    \label{fig:example:clusters}
\end{figure}

Figure \ref{fig:example:clusters} shows some examples of the normal clusters obtained on images from 3 of the scenes in our dataset.

\section{Heavily distorted vanishing-point rectified images}
\label{sec:vp-distortion}
\begin{figure}
    \centering
    \includegraphics[width=0.49\textwidth]{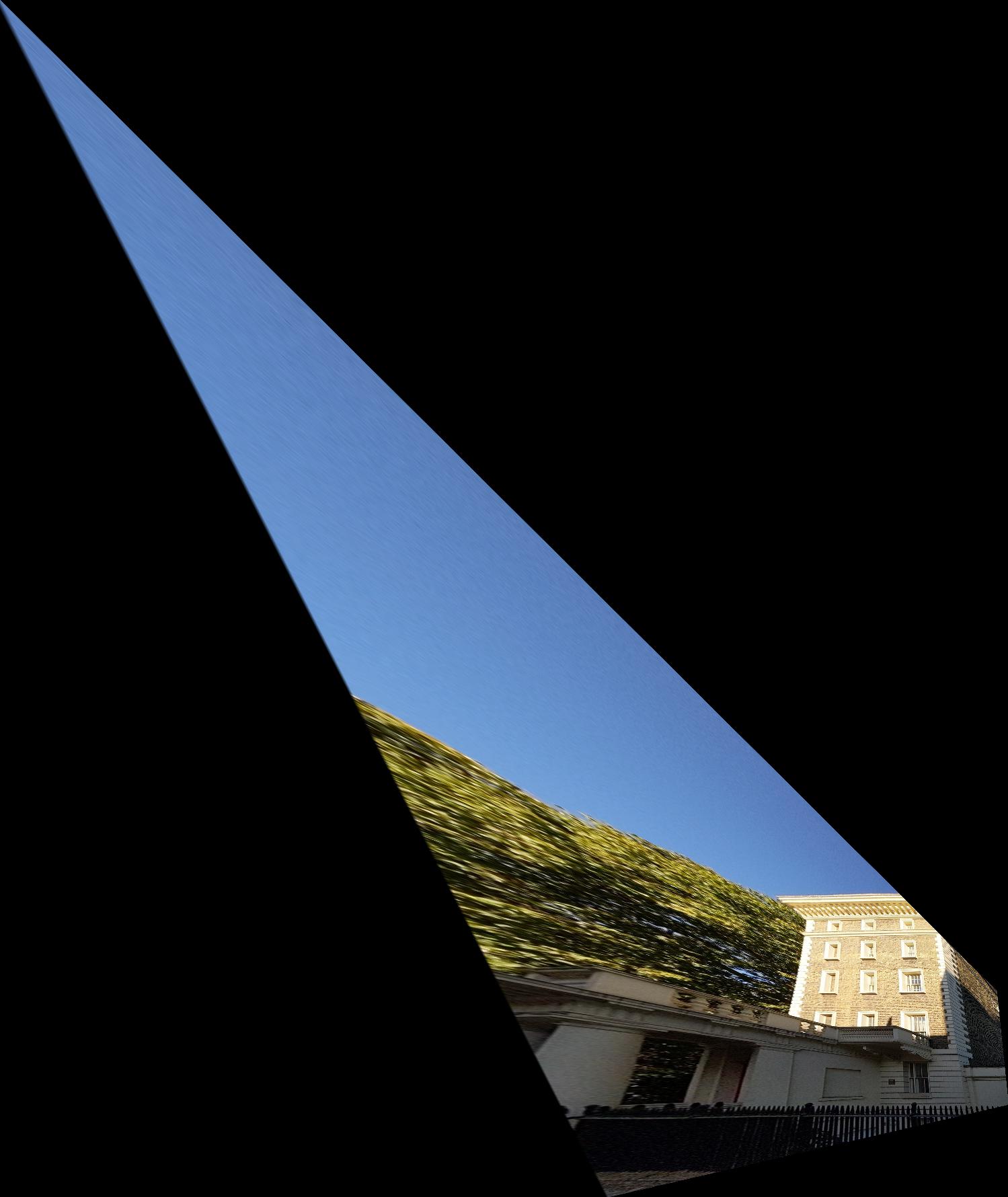}
    \includegraphics[width=0.49\textwidth]{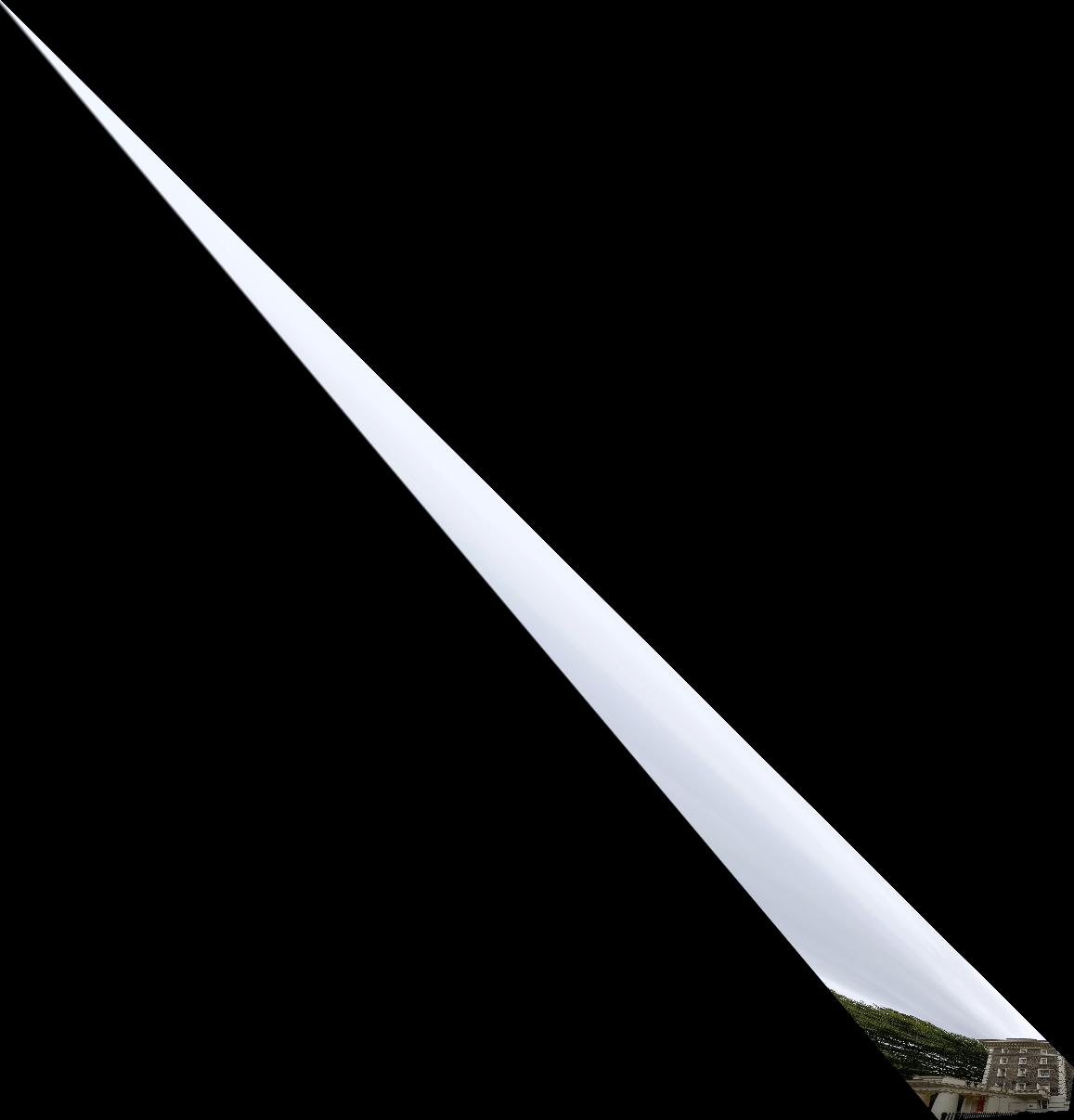} \\
    \includegraphics[width=0.49\textwidth]{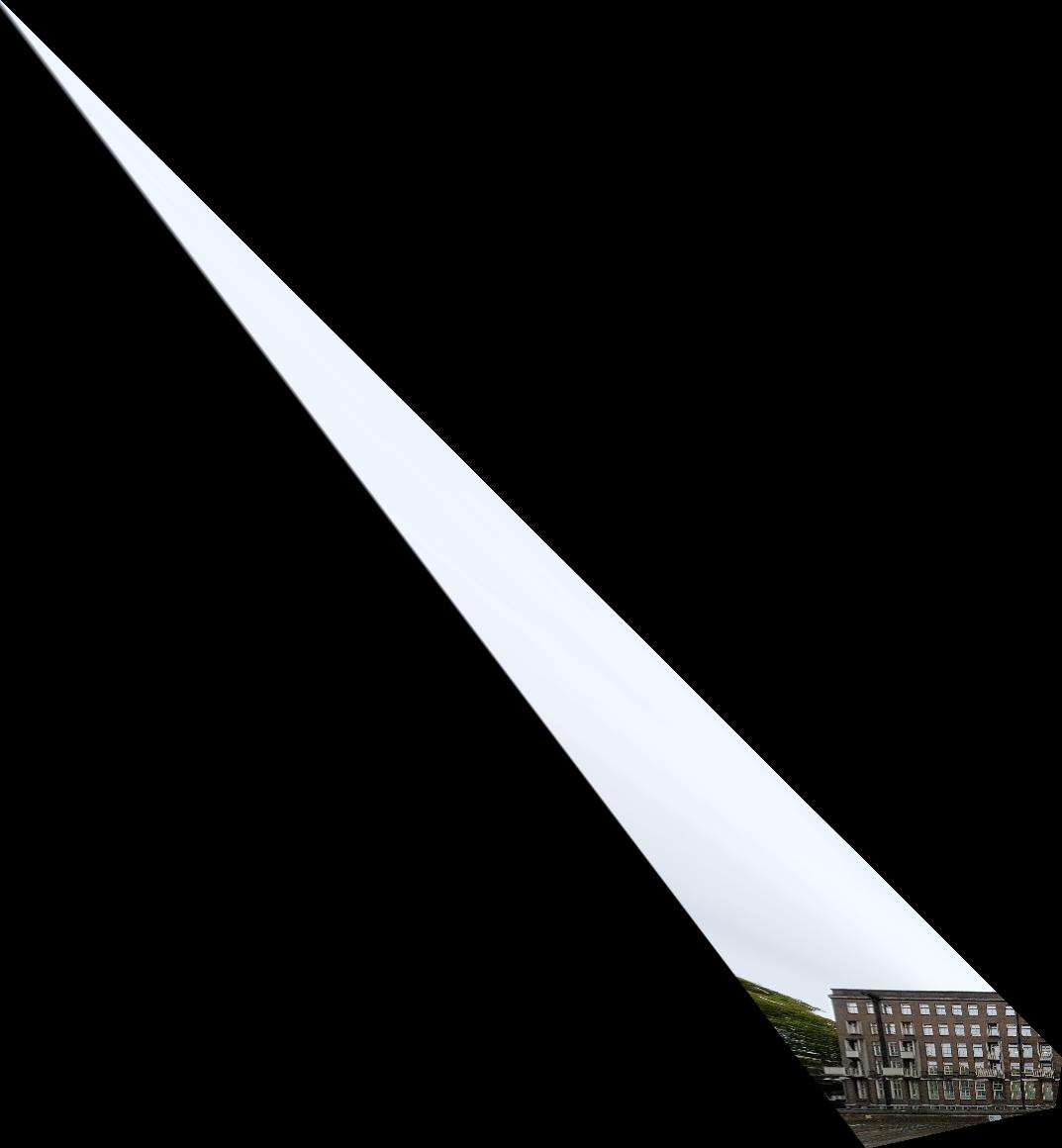}
    \includegraphics[width=0.49\textwidth]{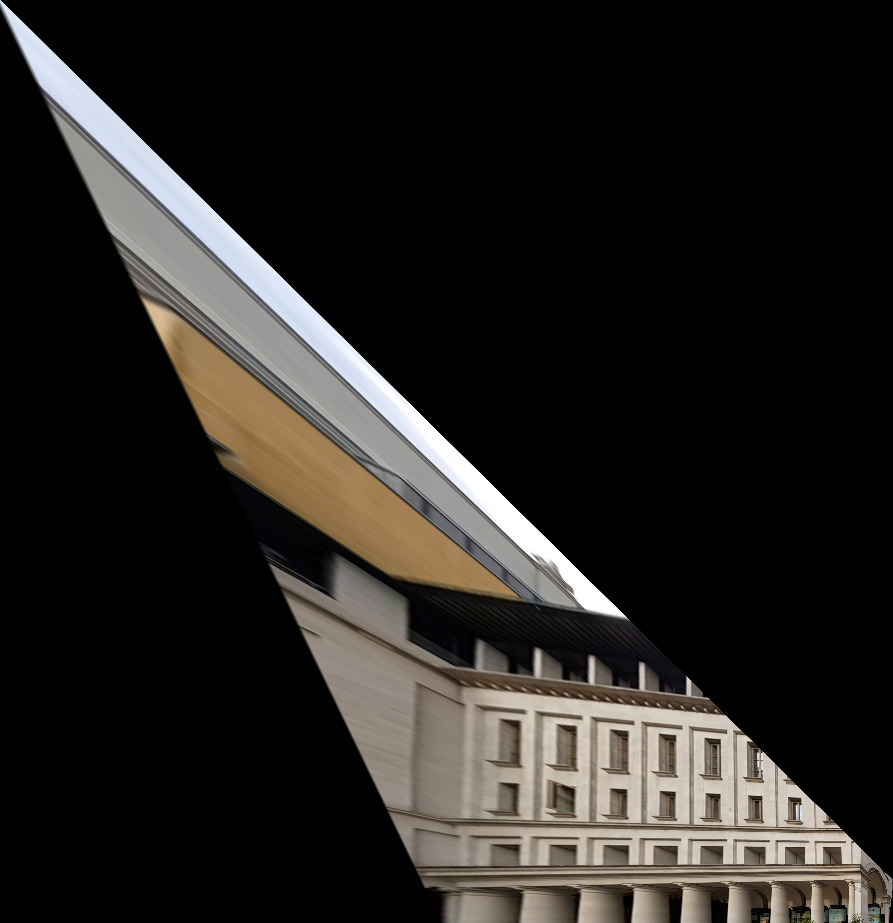}
    \caption{Examples of heavily distorted images that have been rectified using a vanishing point based rectification method. }
    \label{fig:vp-rect-distortion}
\end{figure}

Fig. \ref{fig:vp-rect-distortion} shows heavily distorted images that have been rectified using a vanishing point based rectification method. Since the vanishing point based method does not provide information about which pixels belong to the plane, the entire image is rectified, which can cause strong distortions, and since the entire image is warped, the area of interest may only occupy a small portion of the rectified image.

\section{Experiments on EVD} 
We also ran experiments on three of scenes from the challenging the extreme view dataset (EVD) \cite{mishkin2013two}. The scenes tested were Caf\'e, Dum, Grand. Our method was able to successfully match the Caf\'e scene, but was unable to estimate the homography between the image pairs of the two other scenes. This is likely mainly due to two reasons. First, our method needs the camera intrinsics in order to compute the surface normals from the depthmap, and the dataset does not provide camera calibration information. Secondly, the other scenes contain some non-planar parts, which may cause the estimated plane normals to not be completely accurate. We note that regular feature matching on the original image pairs fails for all three pairs.

\begin{figure}
    \centering
    \includegraphics[width=0.48\textwidth]{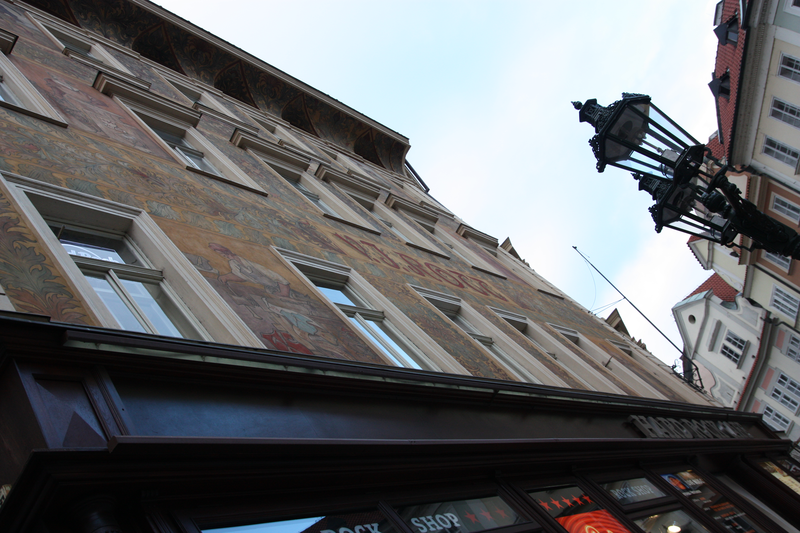}
    \includegraphics[width=0.48\textwidth]{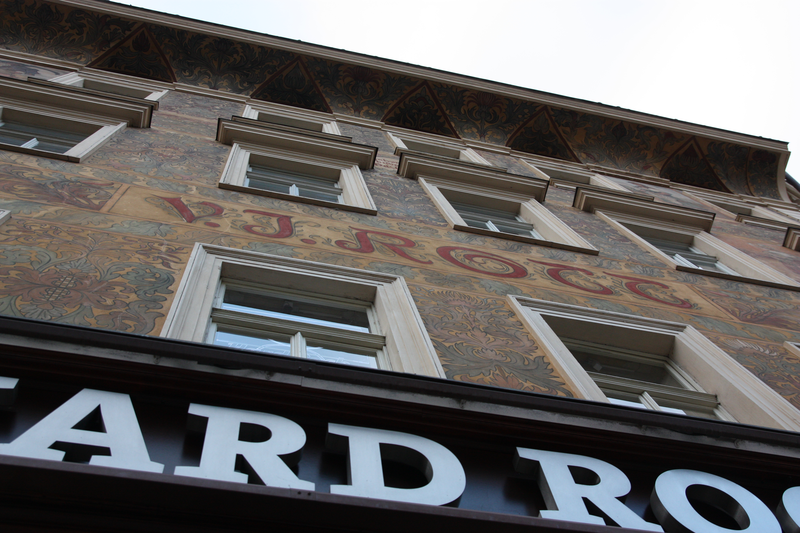} \\
    \hspace*{-4em}\includegraphics[width=1.2\textwidth]{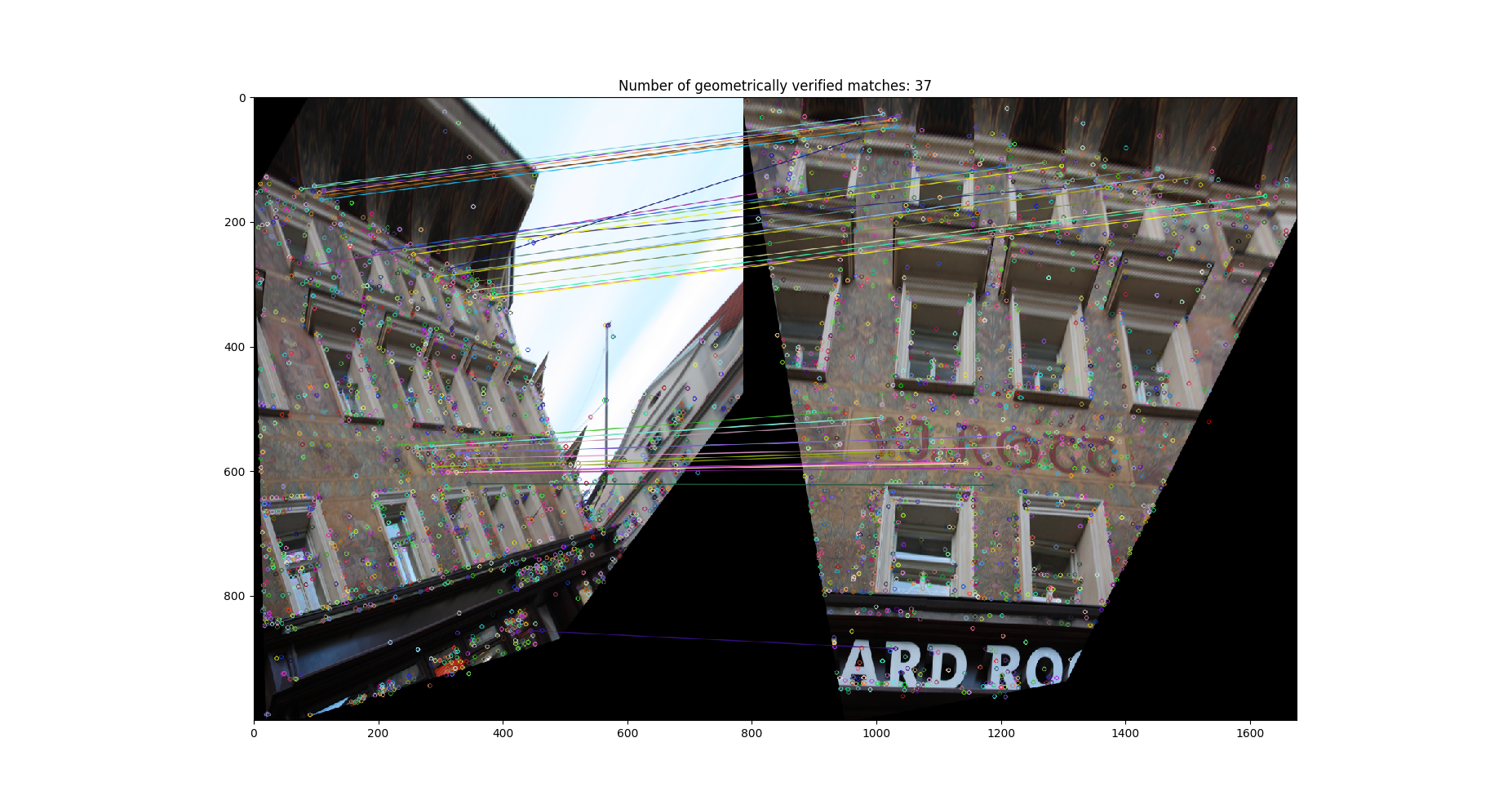}
    \caption{Results on the Caf\'e scene in the EVD dataset. Top row: Original images. Bottom row: Geometrically consistent matches between the rectified patches. }
    \label{fig:cafe}
\end{figure}

Fig. \ref{fig:cafe} shows the results on the Caf\'e scene. 

\clearpage
%
%
\bibliographystyle{splncs04}
\bibliography{main_bib}